%% file: main.tex
\documentclass[conference]{IEEEtran}
\IEEEoverridecommandlockouts
\usepackage{cite}
\usepackage{float}
\usepackage{amsmath,amssymb,amsfonts}
\usepackage{algorithm}
\usepackage{algpseudocode}
\usepackage[pagebackref=true,breaklinks=true,colorlinks,bookmarks=false]{hyperref} 
\usepackage[table]{xcolor}
\usepackage{colortbl}
\usepackage{subcaption}
\usepackage{times}
\usepackage{makecell}
\usepackage{multirow}
\usepackage{url}            % simple URL typesetting
\usepackage{booktabs}       % professional-quality tables
\usepackage{amsfonts}       % blackboard math symbols
\usepackage{nicefrac}       % compact symbols for 1/2, etc.
\usepackage{microtype}      % microtypography
\usepackage{amsmath}
\usepackage{bbding}
\usepackage{float}
\usepackage{caption}
\usepackage{subcaption}
\usepackage{wrapfig}

\usepackage{graphicx}
\usepackage{textcomp}
\usepackage{xcolor}
\usepackage{listings}
\usepackage{adjustbox}
\usepackage{helvet}
\def\BibTeX{{\rm B\kern-.05em{\sc i\kern-.025em b}\kern-.08em
    T\kern-.1667em\lower.7ex\hbox{E}\kern-.125emX}}
    
\begin{document}

\title{Exploiting Inter-Layer Expert Affinity for Accelerating Mixture-of-Experts Model Inference}

\author{
    \IEEEauthorblockN{
        Jinghan Yao,
        Quentin Anthony,
        Aamir Shafi,
        Hari Subramoni,
        Dhabaleswar K. (DK) Panda
    }
    \IEEEauthorblockA{
        \textit{Department of Computer Science and Engineering} \\
        \textit{The Ohio State University}\\
        Columbus, OH, U.S. \\
        \{yao.877, anthony.301, shafi.16, subramoni.1\}@osu.edu, panda@cse.ohio-state.edu
    }
    \thanks{*This research is supported in part by NSF grants \#1818253, \#1854828, \#1931537, \#2007991, \#2018627, \#2112606, \#2311830, \#2312927, and XRAC grant \#NCR-130002.}
}

%\linespread{0.95}

\maketitle

\input{0-abstract}
%\maketitle
\input{1-introduction}
\input{2-background}
\input{3-motivation}
\input{4-design}
\input{5-experiments}
\input{7-relatedwork}
\input{6-contributions}

\bibliographystyle{IEEEtran}
\bibliography{main}

\input{appendix}

\end{document}

%% file: 0-abstract.tex
\begin{abstract}

In the realm of large language models (LLMs) like the Generative Pre-trained Transformer (GPT), the Mixture of Experts (MoE) paradigm has emerged as a powerful technique for enhancing model expressiveness and accuracy. However, the deployment of GPT MoE models for parallel inference on distributed systems presents significant challenges, primarily due to the extensive Alltoall communication required for expert routing and aggregation. This communication bottleneck exacerbates the already complex computational landscape, hindering the efficient utilization of high-performance computing resources. In this paper, we propose a lightweight optimization technique called ExFlow, to largely accelerate the inference of these MoE models. We take a new perspective on alleviating the communication overhead by exploiting the inter-layer expert affinity. Unlike previous methods, our solution can be directly applied to pre-trained MoE models without any fine-tuning or accuracy degradation. By proposing a context-coherent expert parallelism on distributed systems, our ExFlow design only uses one Alltoall communication to deliver the same functionality while previous methods all require two Alltoalls. By carefully examining the conditional probability in tokens' routing across multiple layers, we proved that pre-trained GPT MoE models implicitly exhibit a strong inter-layer expert affinity. We then design an efficient integer programming model to precisely capture such features and show that by properly placing the experts on corresponding GPUs, we can reduce up to 67\% of tokens' cross-GPU routing latency on various hardware configurations and topologies. Our solution beats the cutting-edge Deepspeed-MoE in GPT MoE models with experts from 8 to 64, with up to \textbf{2.2x} improvement in inference throughput. To the best of our knowledge, this is the first work in leveraging inter-layer expert affinity to accelerate the inference of GPT MoE models. We further provide a detailed study of how the model implicitly acquires this expert affinity at the very early training stage and how this affinity evolves and stabilizes during training.
\end{abstract}

%% file: 1-introduction.tex
% \vspace{-1.0ex}
\section{Introduction}
\label{sec:introduction}
% \vspace{-1.0ex}

\begin{figure}[tbp]
    \centering
    \includegraphics[width=0.47\textwidth]{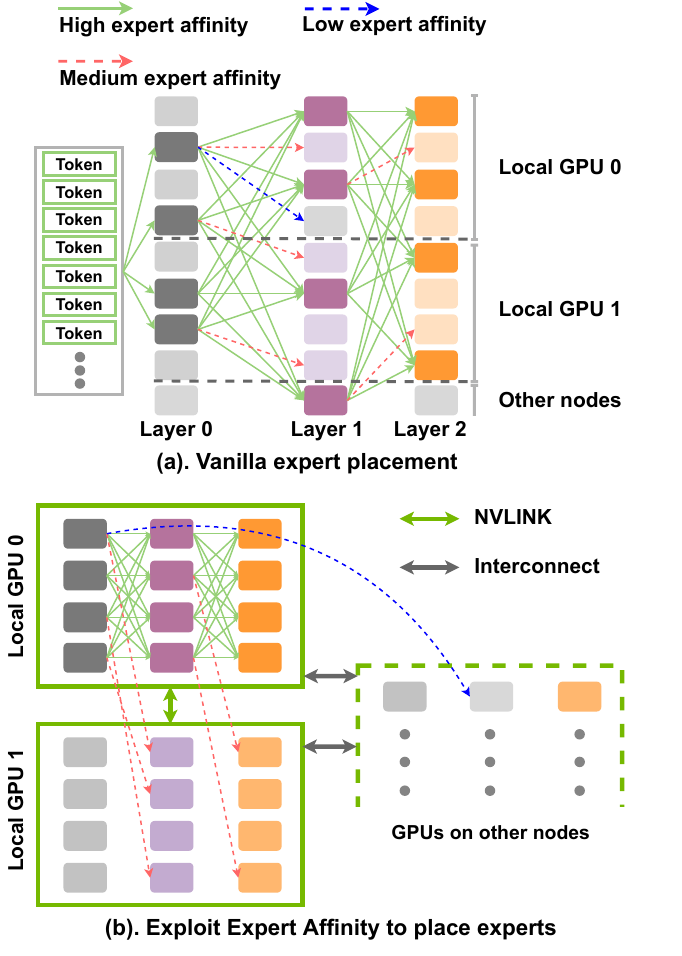}
    \caption{Given a pre-trained MoE model, (a) vanilla placement strategy causes intensive cross-GPU communication. (b) leveraging inter-layer expert affinity can avoid unnecessary Alltoall communication.}
    \label{fig:intro}
    \vspace{-3.5ex}
\end{figure}
In the evolving landscape of artificial intelligence (AI) and deep learning (DL), the Mixture of Experts (MoE)~\cite{fedus2022switch, shazeer2017outrageously, zhou2022mixture, masoudnia2014mixture, he2021fastmoe, artetxe2021efficient} paradigm has emerged as a pivotal technique, bolstering the efficiency and adaptability of models. MoE operates on the principle of distributing tasks among specialized experts within a broader model architecture, dynamically routing the input to the most adept expert based on the context. While MoE is a domain-agnostic technique that has achieved success in domains such as vision \cite{shen2023scaling, riquelme2021scaling}, MoE has been particularly instrumental for large language models (LLMs)~\cite{radford2019language, brown2020language, openai2023gpt4} in scaling language modeling capabilities while constraining computational costs. While being powerful in scaling the capacity of large language models, it usually requires special parallelism strategies to alleviate the memory requirement, as accommodating all experts on a single GPU is infeasible due to each expert being a de facto large feed-forward network~(FFN). Modern LLMs usually deploy multiple Mixture-of-Experts layers, previously proposed expert parallelism allows each GPU to load only a few experts per MoE layer, given its global rank. However, by introducing more GPUs to hold different experts, the overhead of communication becomes non-trivial, especially for the latency-sensitive inference stage. 

\begin{figure*}[t]
  \vspace{-1em}
  \centering
  \begin{subfigure}[b]{0.24\textwidth}
    \centering
    \includegraphics[width=\linewidth]{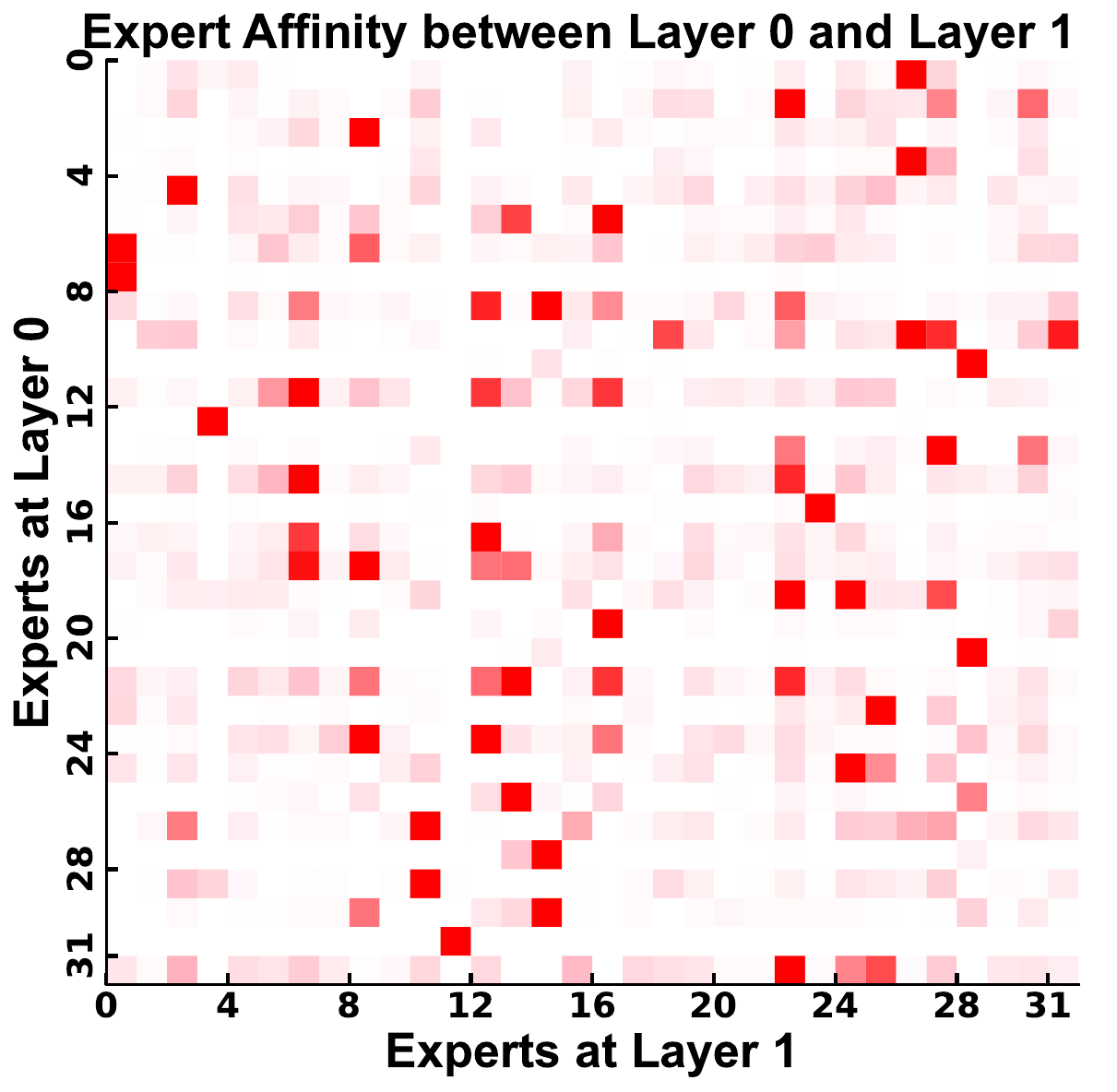}
    \caption{Layer 0 and layer 1}
    \label{fig:prop_1node}
  \end{subfigure}
  \begin{subfigure}[b]{0.24\textwidth}
    \centering
    \includegraphics[width=\linewidth]{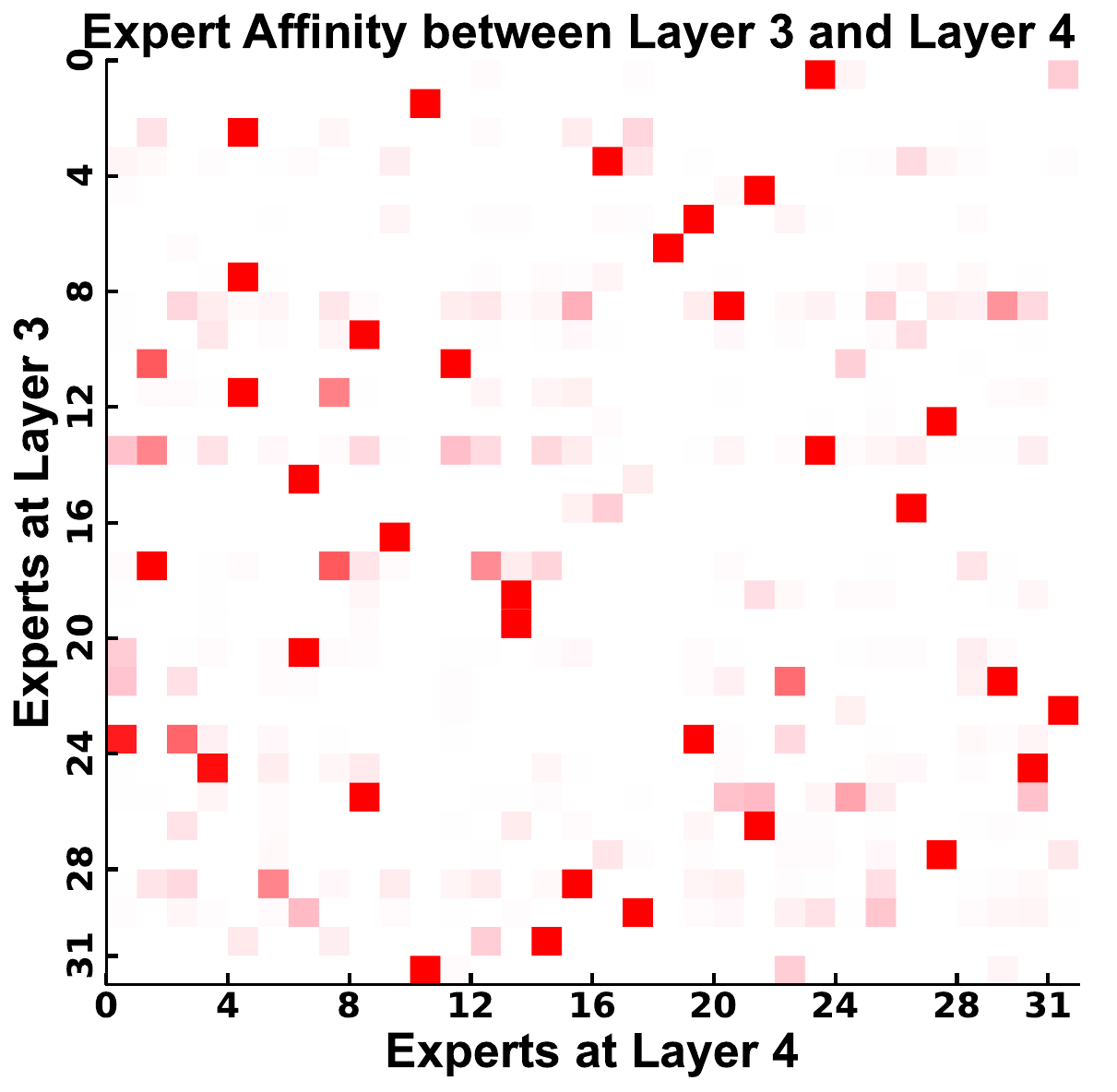}
    \caption{Layer 3 and layer 4}
    \label{fig:prop_2node}
  \end{subfigure}
  \begin{subfigure}[b]{0.24\textwidth}
    \centering
    \includegraphics[width=\linewidth]{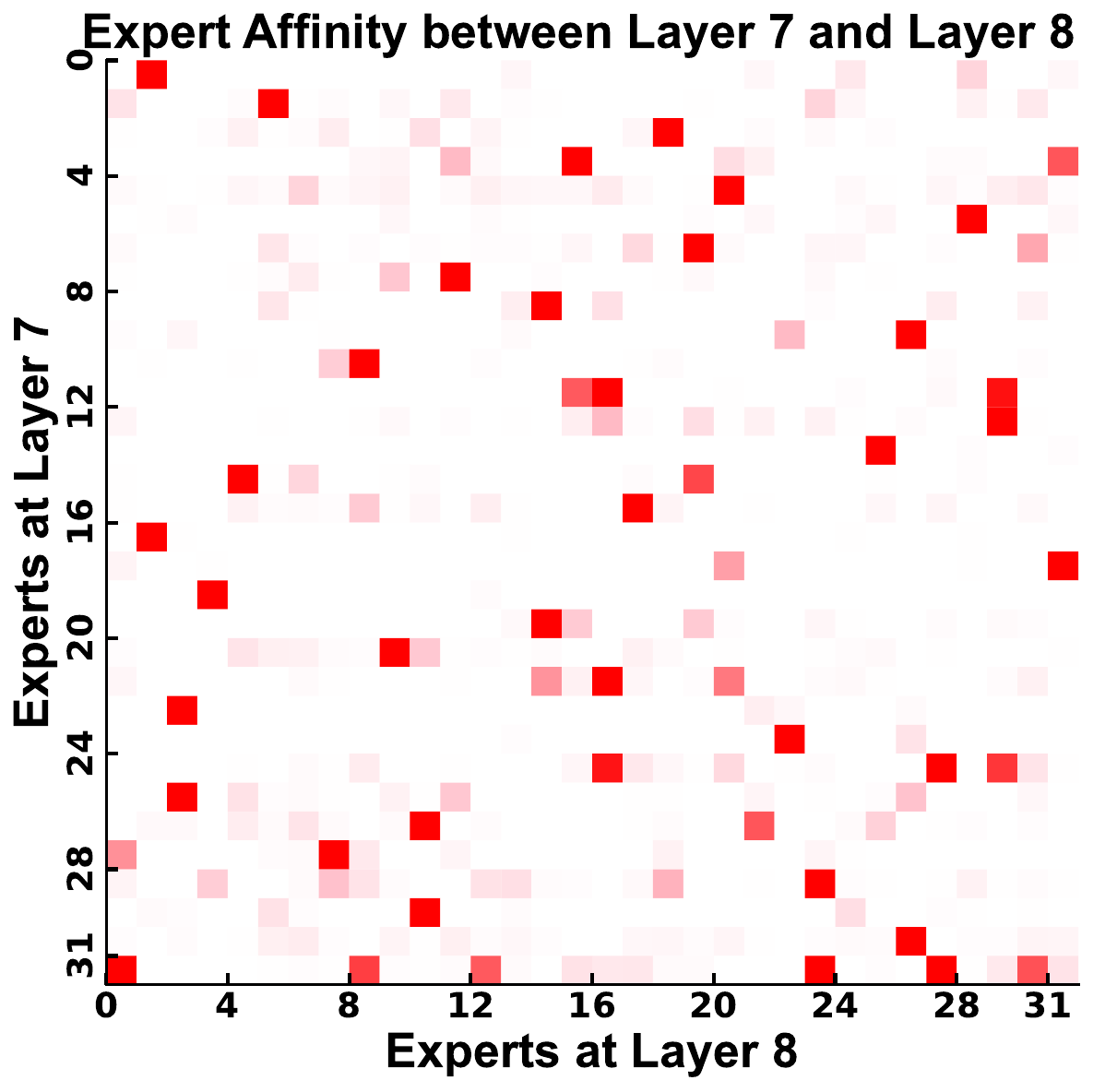}
    \caption{Layer 7 and layer 8}
    \label{fig:prop_4node}
  \end{subfigure}
  \begin{subfigure}[b]{0.243\textwidth}
    \centering
    \includegraphics[width=\linewidth]{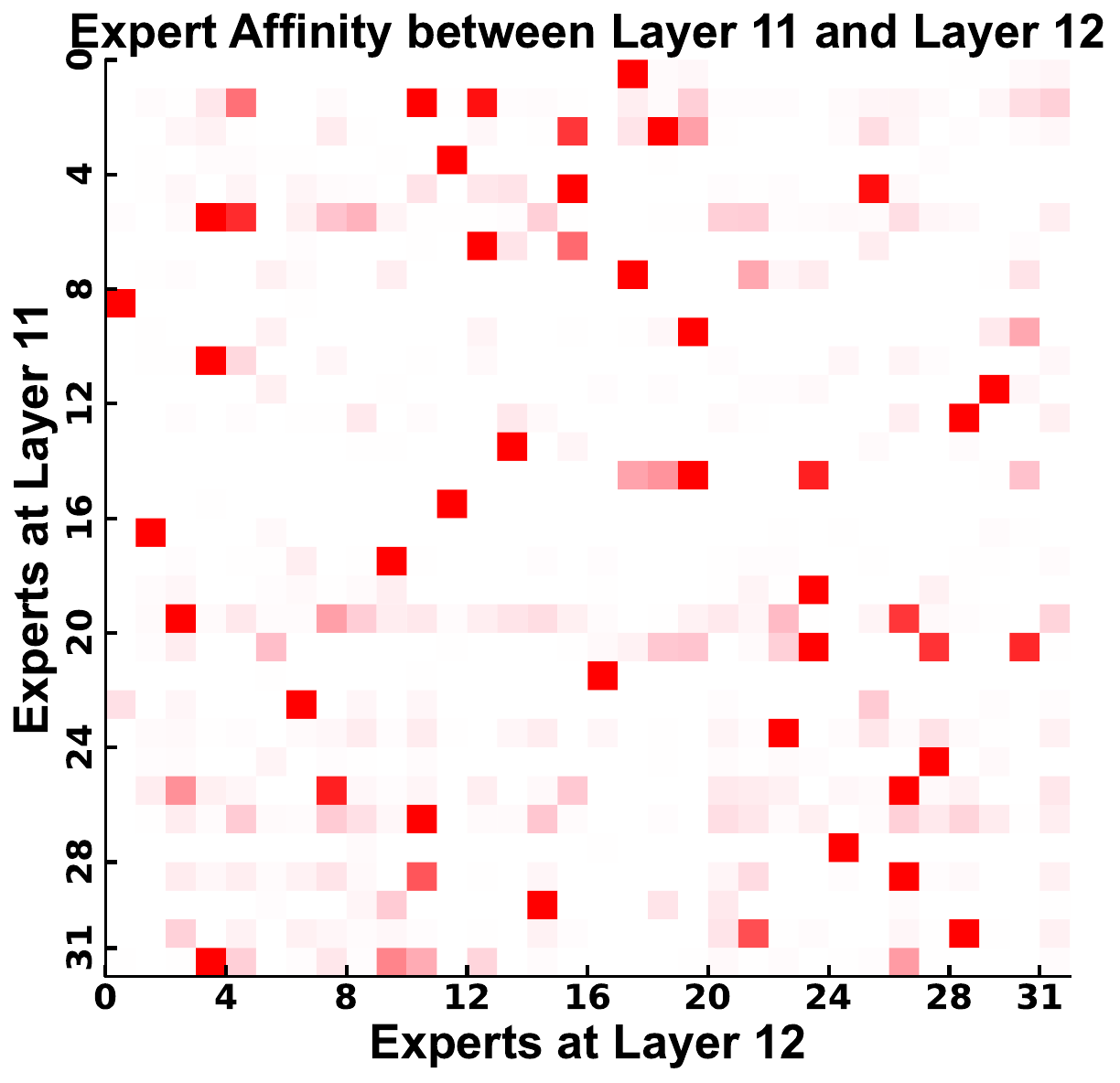}
    \caption{Layer 11 and layer 12}
    \label{fig:prop_8node}
  \end{subfigure}
  \caption{Heatmaps illustrating the distribution of inter-layer expert routing preference. Color intensity represents the magnitude of the likelihood, with white signifying low values and red indicating high values. We measure the conditional probability of expert routing in different parts of a pre-trained GPT model with 12 MoE layers, and 32 experts per layer. For each row, we can observe only a few columns are red, indicating a strong affinity.}
  \label{fig:cond_prob}
  \vspace{-1em}
\end{figure*}

\subsection{Problem Statement}
Current MoE~\cite{rajbhandari2022deepspeed, he2022fastermoe, chen2022ta, artetxe2021efficient} working patterns strictly require two Alltoall collectives in every MoE layer, as according to the routing decision made by the gating function, each GPU will first scatter its input to experts on other GPUs, and later gather them back after the computation. Depending on the involved number of GPUs, Alltoall collectives will introduce a significant overhead. Existing solutions~\cite{he2022fastermoe, chen2022ta} introduce topology-aware loss during training, trying to let the gating function route more tokens to local GPUs with less communication latency, however, while it can accelerate the training, this heuristic constraint impedes the model's performance and becomes invalid during the inference stage once the hardware topology is changed. Given the resource-intensive nature of LLM training and the varied inference scenarios, a universally applicable, communication-efficient MoE routing design remains a pressing requirement.

\subsection{Motivation}
In a Mixture of Experts (MoE) layer, each expert model specializes in a distinct domain of knowledge~\cite{hoefler2021sparsity, yosinski2014transferable}. Modern MoE models usually stack multiple such MoE layers so that at each layer, the input will be routed to one or a few experts. The domain knowledge that each expert is responsible for might vary on different models and trainings. However, for a pre-trained MoE model, we are curious about whether there exist some correlations between the expert selection across different MoE layers. In other words, for a pre-trained MoE model with multiple experts per layer, given an input token, if we know that it is routed to a specific expert at layer $i$, what is the likelihood of the token's routing destination at the next layers? \textbf{Will it be purely stochastic? Or will certain experts exhibit a higher probability of being selected as the next destination?}

Fig~\ref{fig:cond_prob} shows the heatmap of the routing preference on a pre-trained GPT~\cite{radford2018improving, child2019generating} MoE model. The model consists of 12 MoE layers, and each layer has 32 experts, refer to \ref{tab:models} for more details. We select four pairs of consecutive layers and trace the expert selection of tokens in these layers. The Y-axis depicts the expert on the previous layer, and X-axis depicts the expert on the following layer. The red block on coordinate $(x,y)$ represents the conditional probability of tokens being routed to $expert_y$ at $layer_i$ then being routed to $expert_x$ at $layer_{i+1}$. The more intense the color, the higher the conditional probability it is. Telling by all four heatmaps, we can clearly observe that expert selection is not random and routing decisions in previous layers will largely affect the later layer's routing decisions, and this is true for any layers in the model. Therefore, we define such a conditional probability in expert selections across different layers as \textbf{expert affinity}. \cite{li2023accelerating, yi2023edgemoe} previously observed this phenomenon, yet it has not been extensively researched, prompting our in-depth exploration and subsequent optimization proposal.

\subsection{Proposed Solutions}
In this paper, we introduce a new perspective on optimizing the MoE Alltoall patterns by going beyond individual MoE layers and exploiting the inter-layer experts' affinity. With a careful examination of the implicit data parallelism in current expert parallelism, we propose a context coherence design where every GPU holds the contexts of all tokens in processing, which then allows us to directly cut down by half of the Alltoall operations in each MoE layer. Furthermore, by exploiting expert affinity across multiple MoE layers, we can reduce up to 40\% of data exchange in the remaining Alltoall communication, without creating any replicas of additional experts that do not belong to the current GPU rank. Our optimization strategy is adapted to various topologies of compute nodes, leveraging the hierarchical bandwidth of GPU memory copies, intra-node intra-node NVLINK~\cite{nvidia2022nvlink}, and inter-node networks. Our solution can be quickly applied to various pre-trained GPT MoE models without any prior re-training or fine-tuning on the model, and is guaranteed to bring benefits regardless of the number of experts that can be stored within a single GPU's memory capacity.

% We start with a brief demonstration of expert affinity. 

To the best of our knowledge, this is the first work in exploring inter-layer expert affinity to accelerate pre-trained GPT MoE inference. Therefore, upon thorough examination of the prevailing paradigms in expert parallelism, we enumerate our contributions as follows:

% Current expert parallelism is indeed a combination of data parallelism and model parallelism. For example, a typical GPT MoE-32 model with an expert parallel degree of 8 is actually a DP-8 MP-8 parallel, meaning eight GPUs, each holding its own input data, and 4 experts per layer. Note that other parts of the model are shared among all GPUs, e.g. the attention module, normalization layers, etc. Such a parallelism design forces each GPU to hold the same tokens throughout the model in every iteration, because each token needs to attend to its context so that it can perform the crucial attention computation, as this is the most crucial part of large language models. The context could be the input prompt words or previously generated tokens, however, these context tokens are stored locally on each GPU, and they are strictly bonded to the current running tokens on that GPU, as they are in the same context. The limitation of such parallelism means that after each GPU dispatches its own piece of tokens to experts on other GPUs, it will have to retrieve these tokens when those remote experts finish the computation. And since each MoE layer consists of an attention module and an expert module, and each GPU will dispatch and gather its own tokens, we need two Alltoall operations for every layer.

\begin{enumerate}
    \item We exploit the {\em expert affinity} property that exists implicitly in current pre-trained GPT MoE models by capturing the combined conditional probability of the expert routing decisions across multiple MoE layers.
    \item We thoroughly analyze the computation and communication patterns in existing parallelism strategies, and propose a context-coherent expert parallelism to largely reduce the number of Alltoall collectives in current GPT MoE models.
    \item We design an efficient yet accurate offline algorithm to capture {\em expert affinity} in any pre-trained GPT MoE models by formulating it as an Integer Linear Programming problem, allowing for near-optimal solutions. The derived results inform the expert placement strategy during GPU model loading. 
    \item We propose a novel expert parallelism optimization solution based on context coherence and expert affinity, named ExFlow. It is implementation agnostic and easily applied to any given GPT MoE model. ExFlow can significantly accelerate the inference of these models.
    \item We compare ExFlow with existing advanced MoE inference frameworks on a variety of pre-trained GPT MoE models. For GPT MoE-16/32/64 models, our solution provides up to 56\%/65\%/67\% reduction in Alltoall communication and 120\%/60\%/80\% improvement in inference throughput, respectively.
\end{enumerate}

% \begin{figure}[htbp]
%     \centering
%     \includegraphics[width=0.40\textwidth, page=1]{heatmaps.png}
%     \caption{\textcolor{red}{I will draw a new figure. It takes some time.}}
%     \label{fig:critical_path}
%     \vspace{-1.5ex}
% \end{figure}

%% file: 2-background.tex
\section{Background}
\label{sec:background}

\subsection{Expert Model}
Expert models, particularly in vision and language domains, operate on the principle of distributed specialization, where each expert is responsible for a specific subset of the overall knowledge domain. These experts are typically neural network modules trained to excel in tasks like object recognition, semantic parsing, or sentiment analysis. The overall system employs a gating mechanism to route input data to the most relevant experts, thereby leveraging domain-specific knowledge for improved performance. Expert parallelism is a computational strategy that enables the concurrent execution of these experts, significantly accelerating inference and training. This is crucial for handling large-scale data and complex models, as it allows for the distribution of computational load across multiple hardware accelerators.

Transitioning to large language models (LLMs), the Mixture of Experts (MoE) architecture has been instrumental in scaling their capacity without a linear increase in computational cost. MoE integrates multiple experts into a single model and uses a soft gating mechanism to combine their outputs. However, expert parallelism in MoE necessitates two critical Alltoall operations: 1) routing the input data to appropriate experts, and 2) aggregating the outputs from all experts for the final prediction. While these operations are essential for the model's functionality, they introduce significant communication overhead, especially in distributed inference scenarios involving multiple GPUs. This overhead can become a bottleneck, impeding the scalability and efficiency of the system.

% \begin{table*}[t]
% \centering
% \small % Reduce font size
% \begin{tabular}{c|c|c|c}
% \toprule[2pt]
%       & Deepspeed                            & FasterMoE        & \textbf{Ours} \\ 
% \toprule[1pt]
% \multicolumn{4}{l}{\emph{\textbf{Before inference:}}} \\
% \hline
% Allgather  &    $-$ & $-$   &  $1$                \\ 
% \toprule[1pt]
% \multicolumn{4}{l}{\emph{\textbf{Each layer:}}} \\
% \hline
% Alltoall &    $2$ & $2$   &  $1 $                \\ 
% \toprule[1pt]
% \multicolumn{4}{l}{\emph{\textbf{Each iteration:}}} \\
% \hline
% Alltoall &    $2\cdot L$ & $2\cdot L$   &  $L $                \\ \hline
% Allgather &    $-$ & $-$   &  $1 $                \\ 
% \toprule[2pt]
% \multicolumn{4}{l}{\emph{\textbf{Overall:}}} \\
% \hline
% Alltoall &    $T\cdot 2\cdot L$ & $T\cdot 2\cdot L$   &  $T\cdot L$                \\ \hline
% Allgather &    $-$ & $-$   &  $T + 1$                \\
% \toprule[2pt]
% \end{tabular}
% \caption{Total number of collective operations comparison. $T$ denotes the total number of iterations that each inference will run. $L$ denotes the total number of MoE layers in the model.}
% \label{tab:rm-padding-cnt}
% \end{table*}

\begin{table*}[t]
\centering
\small % Reduce font size
\begin{tabular}{c|c|c|c|c|c|c}
\toprule[2pt]
   & \makecell{Topology \\ Aware}                           & \makecell{Additional\\ Topo. loss}    &  Extra Memory  &  \makecell{Forward comm. in\\ Top-1 gating}  &  \makecell{Forward comm. in\\ Top-2 gating}  & \makecell{Applicable in \\ Inference} \\ 
\toprule[1pt]
FasterMoE~\cite{he2022fastermoe} & \checkmark & \checkmark &\checkmark & $2\cdot G\cdot N\cdot L\cdot p_{topo}$ & $4\cdot G\cdot N\cdot L\cdot p_{topo}$& \\
TA-MoE~\cite{chen2022ta} & \checkmark & \checkmark & & $2\cdot G\cdot N\cdot L\cdot p_{topo}$ & $4\cdot G\cdot N\cdot L\cdot p_{topo}$& \\
Deepspeed-MoE~\cite{rajbhandari2022deepspeed} & & & & $2\cdot G\cdot N\cdot L\cdot p$ & $4\cdot G\cdot N\cdot L\cdot p$& \checkmark \\
\textbf{ExFlow(Ours)} &\checkmark & & & $G\cdot N\cdot (L\cdot p^{\star}+G)$ & $G\cdot N\cdot (2\cdot L\cdot p^{\star}+G)$& \checkmark\\
\toprule[2pt]
\end{tabular}
\caption{Comparison with state-of-the-art GPT MoE optimization methods. $G$ denotes the total number of GPUs in expert parallel group, $N$ denotes the number of tokens per GPU, $L$ denotes the total number of MoE layers in the model. $p$ denotes the actual ratio of tokens that are involved in the Alltoall communication. For FasterMoE~\cite{he2022fastermoe} and TA-MoE~\cite{chen2022ta}, since they adopt topology-aware gating, we use $p_{topo}$. In our method, we exploit inter-layer expert affinity to keep most tokens within their current GPU, which is essentially different from other methods, therefore using $p^{\star}$ to denote the proportion.}
\label{tab:compare}
\end{table*}
\subsection{Gating Strategies and Optimizations}

In Mixture of Experts (MoE) models, the gating function~\cite{fedus2022switch} is a critical component that routes input data to specialized experts, optimizing performance through domain-specific knowledge. GShard~\cite{lepikhin2020gshard} gating employs a softmax-based approach, focusing on computational efficiency but potentially exacerbating communication overhead due to its agnostic approach to hardware topology. On the other hand, topo-aware gating~\cite{he2022fastermoe, chen2022ta} minimizes this overhead by introducing additional loss terms into the training objective, which are sensitive to hardware topology. However, this necessitates retraining the model from scratch, a resource-consuming endeavor especially for large-scale models like GPT.

Table~\ref{tab:compare} compares various cutting-edge MoE frameworks with our proposed design. While topo-aware gating can effectively reduce communication overhead during training, its benefits are not applicable during inference if the hardware topology changes. This limitation poses a significant inconsistency problem for models like GPT, which are often deployed across various hardware configurations. The requirement for retraining with topo-aware gating also adds a layer of computational burden, making it a less practical choice for already trained and deployed models. More importantly, existing optimizations on gating functions still remain within the individual MoE layer, while failing to systematically investigate the overall dataflow across multiple MoE layers in the model.

%% file: 3-motivation.tex
% \section{Motivation and Challenges}
\section{Challenges}
\label{sec:motivation}

% Our \textbf{Motivation} in presenting this paper stems from the observation that, with the advent of large language models (LLMs), Mixture of Experts architectures have gained significant attention for their capacity to improve model performance and efficiency. While the majority of existing research focuses on optimizing the training phase of MoE models, the inference stage remains comparatively underexplored. This gap is particularly critical, as inference is where end-users interact with the model, making latency and throughput pivotal metrics. Efficient inference not only enhances the user experience but also stands to offer substantial cost-savings in high-performance computing environments. Therefore, our paper places a strong emphasis on optimizing MoE models specifically for the inference stage, without requiring any retraining of the model. The objective is to provide a set of best practices and optimization strategies that make pre-trained MoE models not only highly efficient but also widely deployable in real-world, latency-sensitive applications, irrespective of the hardware topology.

\subsection{Data Locality} 
\begin{figure}[h]
    \centering
    \includegraphics[width=0.48\textwidth, page=1]{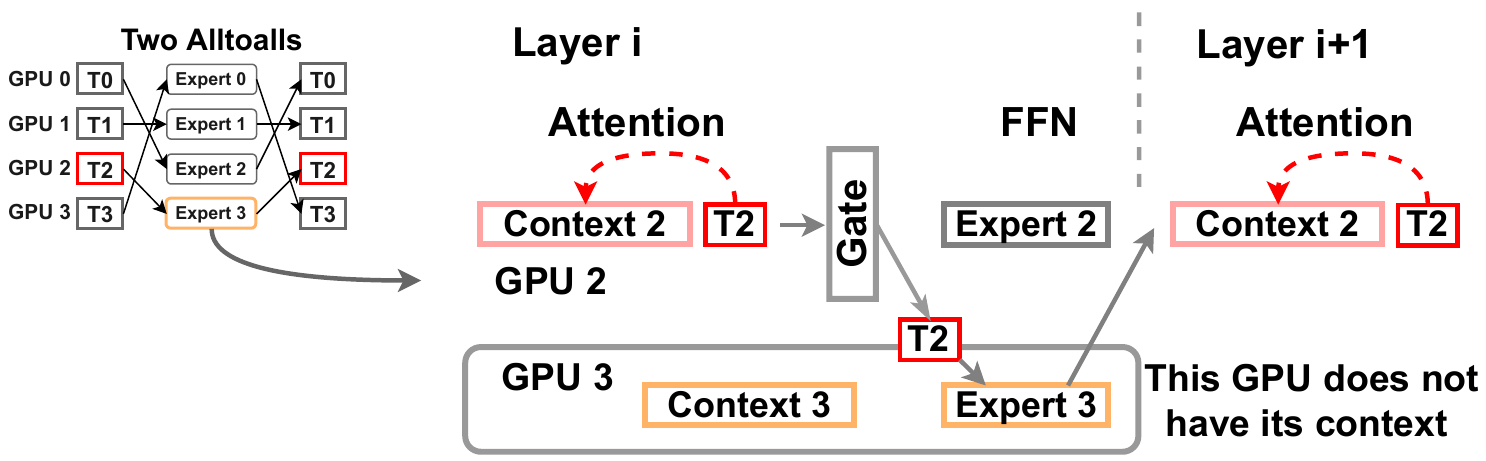}
    \caption{Due to current expert parallelism consisting of data parallel, different GPUs do not share the context of tokens. Therefore, $T_2$ needs to go back to GPU 2 for performing attention in the next layer.}
    \label{fig:context}
    \vspace{-1.5ex}
\end{figure}

As shown in Fig~\ref{fig:context}, prevalent expert parallelism methodologies integrate both data parallelism (DP) and model parallelism (MP)~\cite{shoeybi2019megatron}. In this setup, DP ensures that tokens and their associated contexts, as maintained by individual GPUs, remain distinct. On the other hand, MP ensures each GPU retains exclusive ownership of distinct experts. Therefore, two Alltoall communications are indispensable at every MoE layer. First, in the attention~\cite{vaswani2017attention} calculation, tokens attend to their context on the local GPU, and then a gating function will determine the expert destination of each token. The first Alltoall will route tokens on each GPU to the targeted experts on other GPUs, and this is called token dispatch. Once the first Alltoall is done, each GPU will feed the received tokens to the experts it holds. Note that since experts are essentially FFNs that only perform a non-linear transformation on tokens, they do not require any context information, unlike the previous attention module. However, because multiple MoE layers are stacked, attention modules in the following layer will require tokens to be aligned with the local context, thus at the end of every layer, another Alltoall communication is performed to gather those dispatched tokens. Therefore, in order to remove the second Alltoall operation, we need to overcome the locality constraint of data parallelism, such that tokens can always attend to their corresponding context regardless of their residing GPU.

\subsection{Map Expert Affinity to Hardware Topology} 
Since the goal is to minimize the inference latency of MoE models with expert parallelism, we are interested in reducing the Alltoall communication overhead as much as possible. Therefore, identifying expert affinity in a pre-trained MoE model is our first step, properly mapping this affinity to the underlying hardware is a crucial task. Given a pre-trained MoE model, the inference could happen on a variety of hardware configurations, thus we need a ubiquitous mapping algorithm so that our design can seamlessly adapt to heterogeneous topologies without necessitating any modifications or fine-tuning of the MoE model.

% we need to find out the best expert placement strategy so that once a token is routed to a GPU at the first MoE layer, the remaining experts it will traverse in later layers will likely to also loaded by the current GPU. In the most ideal case, if one GPU can hold all the experts that a token will traverse, then this token requires no communication at all throughout the entire inference. However, our goal is to find a placement strategy such that for any input tokens, a globally minimized routing cost is reached. In other words, on every GPU, when the number of experts held by it remains the same, inter-layer experts should have as high affinity as possible, therefore, tokens that are 
% We thus should treat it as an optimization problem. 

%% file: 4-design.tex
\section{Design of ExFlow}
\label{sec:design}
In this section, we first introduce the context-coherent expert parallelism. We will try to overcome the data locality constraint as mentioned above, as this is crucial when we later exploit expert affinity. Then we will introduce how to model expert affinity in an effective manner, and utilize it to largely accelerate the inference of GPT MoE models.

\subsection{Token Context Coherence in Expert Parallelism}
\begin{figure}[h]
    \centering
    \includegraphics[width=0.47\textwidth, page=1]{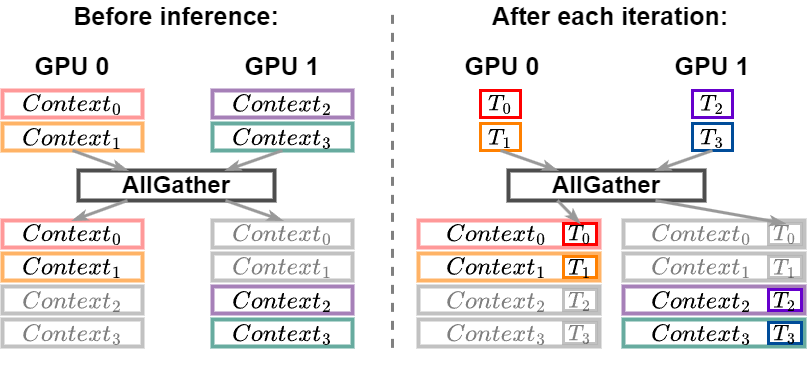}
    \caption{Before inference, we use Allgather to ensure every GPU has all contexts. After each iteration, another Allgather is performed on the newly generated tokens, we then append them to the current contexts for the next iteration.}
    \label{fig:coherence}
    \vspace{-1.5ex}
\end{figure}

Before diving into our design, we would like to revisit the inference pipeline of GPT models. Given a pre-trained GPT model and an inference request, it takes $l$ words per the request as the input, this is usually called \textbf{Prompts}. Prompts are essentially tokens. When the model tries to generate a response for this request, it will refer to prompts for context information. As GPT is indeed a generative model, it will generate one or a few words in each iteration, and append them to the current prompt tokens, which then become the context for generating words in the next iteration. Importantly, once generated, these tokens remain immutable in subsequent iterations, serving purely as context and not undergoing further transformations or updates by the network. For simplicity, we use the term \textbf{context} to represent prompts as well as tokens generated in previous iterations. Suppose we have $n$ GPUs in the data parallel group, to achieve token context coherence across all GPUs in the group, we will focus on both the before-inference and after-iteration parts involved in the overall MoE inference process, as shown in Fig~\ref{fig:coherence}.

\begin{figure*}[htbp]
    \centering
    \includegraphics[width=0.98\textwidth, page=1]{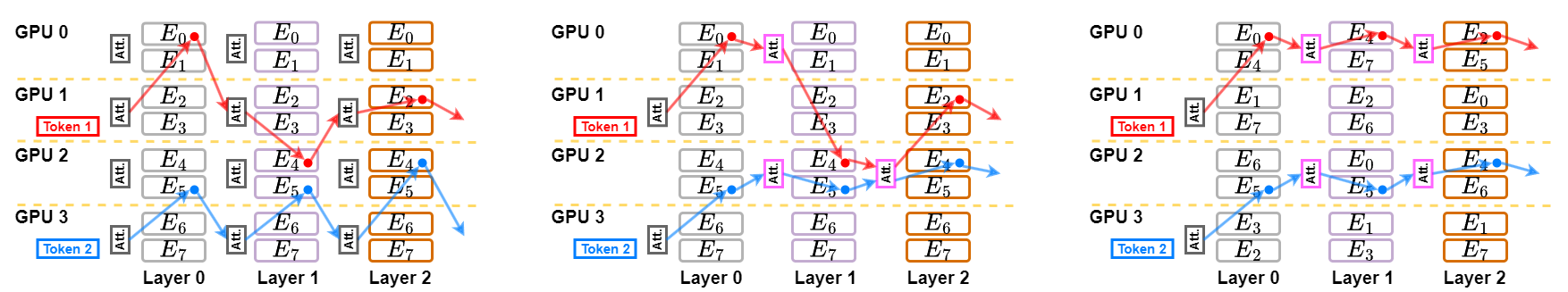}
    \caption{Each GPU has a capacity of 2 experts per layer. (a). Vanilla expert parallelism. Each token needs to come back to its original GPU to perform attention computation with its context. (b). Enabling token context coherence across all GPUs. Tokens do not need to go back to the original GPU to perform attention computation, because they can attend to their context on the local GPU. (c). Exploiting expert affinity to further reduce token communication. The placement of experts on each layer is now following an optimal pattern such that tokens will remain on local GPUs with the maximum probability.}
    \label{fig:compare}
    \vspace{-1.5ex}
\end{figure*}
\textbf{At the start of inference}, GPU $i$ has $g_i$ requests, $i \in \{1,2,\ldots,n\}$, we will first perform an \textbf{AllGather} communication across GPUs where each GPU will broadcast its $g_i$ contexts to all other GPUs. After this, every GPU in the group will have $\sum_{i=1}^{n}g_{i}$ contexts, meaning that all contexts are now coherent across all GPUs. Note that, even though each GPU now has contexts from other GPUs, it will still only generate tokens for its own requests, adhering to data parallelism principles.

\textbf{Upon iteration completion}, each GPU has generated some tokens for its own requests, in order to make sure contexts are still coherent in the next iteration, we need each GPU to broadcast these newly generated tokens to other GPUs. Thus, we also perform an additional \textbf{AllGather} operation across the group. By doing so, at the beginning of a new iteration, every GPU has the up-to-date context of all requests.

\textbf{What are the benefits of ensuring token context coherence?} In Fig~\ref{fig:context}, we mentioned the reason that current expert parallelism strictly requires two Alltoall operations, which is due to the data exclusion as current expert parallelism implicitly exhibits data parallel. Now, as the contexts are coherent and visible on all GPUs for all tokens, a token can perform in-place attention computation with its context, no matter which GPU it is currently on. 

Fig~\ref{fig:compare} shows a 3-layer MoE-8 model running on 4 GPUs, where each GPU holds two experts per layer. We have token 1 on GPU 1 and token 2 on GPU 3. Token 1 will be routed to $E_0$ on layer 0, $E_4$ on layer 1, $E_2$ on layer 2 respectively. Token 2 will be routed to $E_5$ on layer 0, $E_5$ on layer 1, $E_4$ on layer 2 respectively. Fig~\ref{fig:compare}(a) depicts the path that token 1 and token 2 will traverse following the vanilla expert parallel pattern. After each layer, both tokens need to go back to their original GPUs to compute the attention. Notably, for token 2, even if all the experts on its path are on GPU 2, it still has to frequently return to GPU 3 as its context is not coherently visible on GPU 2. For the given example, token 1 requires four cross-GPU communications, and token 2 requires six cross-GPU communications. Fig~\ref{fig:compare}(b) illustrates, however, the path both tokens take when we use context-coherent expert parallelism. For token 1, when it is routed to GPU 0 at layer 0, it finishes the $E_0$ FFN, and performs an in-place attention computation with its context stored on GPU 0. Then, it does not have to come back to GPU 1, rather, it can directly go to $E_4$ at layer 1, which saves 1 cross-GPU communication. For token 2, the improvement is extraordinary, since all the experts on its path are on GPU 2, it only requires one cross-GPU communication on layer 0, after which, all FFNs and attention can be performed in place on GPU 2. Note that, the gating function is shared among all GPUs, so that no matter the token on which GPU, the gating function can always route it to the right expert.

Tab~\ref{tab:compare} shows the overall communication volume in our context-coherent expert parallelism compared to existing methods, such as FasterMoE~\cite{he2022fastermoe}, TA-MoE~\cite{chen2022ta}, and Deepspeed-MoE~\cite{rajbhandari2022deepspeed}. In our design, we cut down half of Alltoalls, while introducing an AllGather at the end of every iteration. We find that as the model has more layers, the overhead of AllGather becomes less significant as it only happens at the last layer.

% Suppose we are running an $L$ layers MoE model with experts distributed on $N$ GPUs, and each GPU is generating $k$ tokens per iteration, In the vanilla expert parallel, the total amount of tokens involved in Alltoall communication is given by~\ref{eq:vanilla}. In context coherence expert parallel, the amount of tokens in both Alltoall and AllGather is given by.

\subsection{Modeling Inter-layer Expert Affinity}
% end-to-end critical path

In this part, we will discuss how to model expert affinity in pre-trained GPT MoE models and how the affinity guides us in reducing Alltoall communications. First, as we are seeking to identify the pattern of expert selection, we need a set of tokens that the model will infer on and we can trace their expert selections at every layer. Here we sample tokens from the Pile~\cite{gao2020pile} dataset to profile the expert routing pattern, a more detailed study on token sampling will be discussed in \ref{sample_tokens}.

In Fig~\ref{fig:cond_prob}, we show the heatmap of consecutive layers' expert selection in a pre-trained GPT 350M MoE-32 model. We achieve this by calculating the conditional probability across experts in consecutive layers. Here we give a mathematical form of expert affinity. Given a pre-trained MoE model with $E$ experts per layer, suppose we have $N$ tokens, denoted by $T_{k}, k \in \{1, 2, \ldots, N\}$, and we denote the $i_{th}$ expert on layer $j$ as $E_{i, j}$. Then the expert affinity between $E_{i,j}$ and $E_{p,j+1}$ can be encapsulated by the following conditional probability:

\begin{gather}
P(E_{p,j+1}|E_{i,j}) = \frac{1}{N}\sum_{k=1}^{N} P(E_{p,j+1}|E_{i,j}, T_k)
\label{eq:affinity}
\end{gather}

For expert $E_{i,j}$, our goal is to find an expert $E_{A^{\star},j+1}$, such that:

\begin{gather}
\frac{1}{n}\sum_{k=1}^{n} P(E_{A^{\star},j+1}|E_{i,j}, T_k) \geq \frac{1}{N}\sum_{k=1}^{N} P(E_{p,j+1}|E_{i,j}, T_k) \notag \\
\text{for all } p \in \{1, 2, \ldots, E\} \text{ with } i \neq A^{\star}
\label{eq:simple}
\end{gather}

We thus claim expert $E_{A^{\star}, j+1}$ is the most affiliated expert with expert $E_{i,j}$. This affinity $P(E_{A^{\star},j+1}|E_{i,j})$ elucidates the likelihood of tokens at $E_{i,j}$ subsequently being routed to $E_{A^{\star},j+1}$. Running a model with 8 experts per layer($E=8$) on eight GPUs, for example, will result in each GPU holding 1 expert per layer. In essence, for any two consecutive layers, we have eight pairs of experts. Strategically placing these affiliated experts on identical GPUs ensures that a token, once routed to a GPU, exhibits a high propensity to remain on that GPU, given that its most affiliated experts in subsequent layers are also resident on the same GPU. However, a challenge arises when formula~\ref{eq:simple} deduces $E_{A^{\star},j+1}$ for multiple experts from layer $j$, leading to repetition. This necessitates a comprehensive strategy to ascertain the globally optimal expert affinity.

Furthermore, when GPUs have a larger capacity, meaning that each GPU can hold more than one expert per layer, as shown in Fig~\ref{fig:compare}, the search space becomes much larger. Given the capacity $C_1$ of a single GPU, denoting the number of experts it will hold per layer, the previous problem changes into the following:

Given experts 
\begin{gather}
E_{x_1, j}, E_{x_2, j}, E_{x_3, j}, ..., E_{x_{C_1}, j}, \notag \\
\text{where }x_{1, ..., C_1} \in \{1, 2, \ldots, E\}
\end{gather}

We want to find experts
\begin{gather}
E_{y_1, j+1}, E_{y_2, j+1}, E_{y_3, j+1}, ..., E_{y_{C_1}, j+1}, \notag \\
\text{where }y_{1, ..., C_1} \in \{1, 2, \ldots, E\}
\end{gather}
that maximizes the following combined conditional probability:
\vspace{-1em}
\begin{gather}
\frac{1}{n}\sum_{k=1}^{n}\sum_{p=1}^{C_1}\sum_{q=1}^{C_1} P(E_{y_q,j+1}|E_{x_p,j}, T_k)
\label{eq:complicated}
\end{gather}

Solving this composite objective function ensures tokens routed to experts $x_1, x_2, x_3, ..., x_C$ at layer $j$ exhibit a high propensity to be subsequently routed to $y_1, y_2, y_3, ..., y_C$ at layer $j+1$.

Fig~\ref{fig:compare}(c) shows how the solution to the above problems can guide us in placing experts such that the volume of Alltoall communication can be minimized. For example, if we know that experts $E_0$ and $E_4$ at layer 0 have a high combined affinity to experts $E_4$ and $E_7$ at layer 1, we can place them onto GPU 0. Similarly, we can find experts $E_2$ and $E_5$ at layer 2 have a high affinity to previous experts. We find that with this placement, token 1 only needs 1 Alltoall communication to get to GPU 0, and it will simply perform in-place attention computation as all its related experts in the following layers are on GPU 0.

\subsection{Staged Experts Affinity}
In Fig~\ref{fig:intro}(b), we depict a more complex but practical scenario, where each GPU holds four experts per layer. Since modern clusters leverage NVLINK for intra-node communication, and high-speed interconnect for inter-node. If a token still needs to be routed to an expert outside of its current GPU, we would want that expert to be held at an intra-node GPU (denoted by the red dash line), rather than a GPU that is on another node (denoted by the blue dash line), as performing intra-node communication has much higher bandwidth and lower latency. Therefore, for GPUs on the same node, we would want the experts they hold to also exhibit some extent of affinity to each other. In this case, each expert has two degrees of affinity. The first degree of affinity is with experts on the same GPU, while the second degree of affinity is with experts on the same node. Thus, we can add a constraint to the previous formula~\ref{eq:complicated}, thus experts on one GPU now have the following form of affinity:

\begin{gather}
\frac{1}{N}\sum_{k=1}^{N}\sum_{p=1}^{C_1}\sum_{q=1}^{C_1} P(E_{y_q,j+1}|E_{x_p,j}, T_k) + \quad \quad \quad \quad \notag \\
\quad \quad \quad \quad \frac{1}{N}\sum_{k=1}^{N}\sum_{p=1}^{C_1}\sum_{o=1}^{C_2-C_1} P(E_{y_o,j+1}|E_{x_p,j}, T_k)
\label{eq:hierarchical}
\end{gather}

Note that $C_2$ denotes the per-layer expert capacity of the entire node, and $E_{y_o,j+1}$ represents expert $y_o$ at layer $j+1$ which is held by other intra-node GPUs.

\subsection{Solving Affinity's duality by Integer Programming}
In formula~\ref{eq:simple}, \ref{eq:complicated}, and \ref{eq:hierarchical}, we provide straightforward objective functions to find such a placement of experts that ensures the best expert affinity. However, as we mentioned, these objective functions only stand in the perspective of one GPU. Finding the best expert placement strategy for all GPUs on all nodes is, however, a complicated combinatorial optimization problem.

To circumvent this, we pivot our focus to its associated Lagrange Dual Problem. The idea is to transform our maximization problem into an equivalent minimization problem, which is computationally more tractable. In essence, high affinity implies minimal token re-routing. Thus, the duality emerges from these intertwined objectives: one seeks positive reinforcement through affinity, and the other targets minimization of disruptions to this affinity. To transition to the dual problem, we must establish the relationship between maximizing this aggregate affinity and minimizing token re-routing costs. This naturally leads us to consider the disruptions to affinity, which can be quantified as token re-routing costs between GPUs.

Given our objective to minimize token re-routing, we form the dual function $g(\lambda)$:
\begin{equation}
g(\lambda, E_{i,j}) = \inf_{E_{p,j+1}} \left[ P(E_{p,j+1} | E_{i,j}) - \lambda G(E_{p,j+1}, E_{i,j}) \right] 
\label{eq:dual_function}
\end{equation}
Where $G(E_{p,j+1}, E_{i,j})$ represents the cost associated with re-routing tokens due to expert selections on disparate GPUs, and $\lambda$ serves as a regularization term to balance affinity and token re-routing.

% \textbf{Proof:} If both the primal (maximizing affinity) and the dual (minimizing cross-GPU traversal) are convex problems, then strong duality holds, implying that the solutions to the primal and dual problems align. Our primal problem is inherently convex, aggregating probabilities. Hence, by minimizing cross-GPU routing in the dual, we are indirectly maximizing affinity.

Since inter-node communication always has the highest latency and lowest bandwidth, our first priority is to reduce the amount of inter-node routing. Fig~\ref{fig:intro} (b) shows an example of the staged expert affinity, where green arrows denote high expert affinity, red ones denote medium level of affinity, and blue ones denote the lowest expert affinity. Our goal is to keep high-affinity experts inside single GPUs, and keep medium-affinity experts inside single nodes. In \textbf{stage 1}, we will reduce the inter-node routing as much as possible, and in \textbf{stage 2}, we will minimize the intra-node routing based on stage 1 results.
Therefore, we propose a coefficient-free objective function following a top-down optimization strategy.

Formula~\ref{eq:obj} is the objective function, where $x_{i, j}^p$ is a binary variable, denoting whether $E_{i,j}$ is held by GPU $p$, $R_{k, j}$ is a binary variable, equaling to one denotes that for a token $k$ at layer $j$, it will be routed to an expert outside of current node(or GPU). Note that, here we apply the exactly identical object function to both stage 1 and 2 (as mentioned above). In stage 1, $R_{k, j}=1$ represents an inter-node routing, while in stage 2, it represents an intra-node routing. For constraints, we need all nodes(or GPUs) to be load-balanced, meaning that for every layer, each node(or GPU) should hold the same number of experts, which is defined by formula~\ref{eq:load_balance}, where $E$ denotes the total number of experts per layer, and $P$ denotes the total number of nodes(or GPUs). Furthermore, for completeness, we need formula~\ref{eq:exclu} to make sure each expert is exclusively held by one node(or GPU). Formula~\ref{eq:cross1} and \ref{eq:cross2} are the essential inequities that map the placement of experts to whether a specific routing is cross-node(or GPU) or not. After solving the above integer linear programming problem, variable $x_{i, j}^p$ in the solution will be directly used as the expert placement strategy when loading the MoE model to GPUs.

\begin{align}
\text{Minimize} \quad & \sum_{k=1}^{N}\sum_{j=1}^{L-1} R_{k, j} \label{eq:obj}\\
\text{Variable} \quad & x_{i, j}^{p} \in \{0, 1\}, R_{k,j} \in \{0, 1\}\notag \\
\text{Subject to} \quad &\sum_{i=1}^{E}x_{i, j}^{p}=\frac{E}{P}, \notag \\ 
&\forall j \in \{1,2,\ldots,L\}, p \in \{1,2,\ldots,P\} \label{eq:load_balance}\\
&\sum_{p=1}^{P}x_{i, j}^{p}=1, \notag \\ 
&\forall j \in \{1,2,\ldots,L\}, i \in \{1,2,\ldots,E\} \label{eq:exclu}\\
&  R_{k,j} \geq x_{i, j}^p - x_{i, j+1}^p\label{eq:cross1}  \\
&  R_{k,j} \geq x_{i, j+1}^p - x_{i, j}^p\label{eq:cross2} 
\end{align}

\subsection{Insensitivity of Expert Affinity}
In order to precisely capture the inter-layer expert affinity in a pre-trained MoE model, we need to use enough tokens and trace their routing decision at each MoE layer. As the goal is to accelerate the inference, we also need to investigate whether expert affinity is insensitive to the distribution of the dataset. The reason is that we cannot predict the actual tokens and their contexts when the model is serving requests from users, therefore, the expert affinity we learned from the offline dataset must remain effective during online inference. We will further analyze this in the experiment.

%% file: 5-experiments.tex
\section{Experimental Evaluation}
\label{sec:experiments}

\subsection{Setups}
\textbf{Hardware: } We conduct all experiments on Wilkes3 Ampere GPU cluster, where each node has 2 AMD EPYC 7763 64-Core processors, and 4 NVIDIA A100-SXM4-80GB GPUs, connected by NVLINK. For inter-node, it is equipped with dual-rail Mellanox HDR200 InfiniBand interconnect.

\textbf{Models: } We use the Deepspeed-Megatron~\cite{megatron_deepspeed, nvidia2021deepspeed} library for pre-training. During inference, all models are with Top-1 gating and variable token capacity.\footnote{Code will be available at \href{https://github.com/YJHMITWEB/ExFlow}{https://github.com/YJHMITWEB/ExFlow.}}
\begin{table}[h]
\centering
% \fontfamily{phv}\selectfont
\small % Reduce font size
\renewcommand{\arraystretch}{1.5}
\begin{tabular}{c|c|c|c|c}
\toprule[1pt]
  Model & Base                         & Experts   &  Layers   & D  \\ 
\hline
\multirow{4}{*}{MoE GPT-M} & \multirow{4}{*}{350M}  & 8 & \multirow{4}{*}{24} & \multirow{4}{*}{1024} \\
 &  & 16 & &  \\
 & & 32& & \\
 & & 64& & \\
\toprule[1pt]
 \multirow{2}{*}{MoE GPT-M}& 470M& \multirow{2}{*}{32}& 32 & \multirow{2}{*}{1024}\\
 & 590M& & 40 & \\
\toprule[1pt]
MoE GPT-XL & 1.3B& 16& 24 & 2048 \\
\toprule[1pt]
\end{tabular}
\caption{List of MoE models we used for experiments.}
% \fontfamily{\rmdefault}\selectfont
\renewcommand{\arraystretch}{1}
\label{tab:models}
\vspace{-1em}
\end{table}

\textbf{Datasets: } \label{dataset}We split the Pile~\cite{gao2020pile} dataset into a training set and an evaluation set. The training set is used to train the model, during the training, we record tokens' expert routing decisions at every layer. We solve the objective function~\ref{eq:obj} based on the tracing logs and then determine the expert placement strategy. Then, we load the model onto GPUs following the placement strategy and benchmark the performance on the evaluation set.

\begin{figure}[h]
\vspace{-1em}
  \centering
  \begin{subfigure}[b]{0.49\textwidth}
    \includegraphics[width=\linewidth]{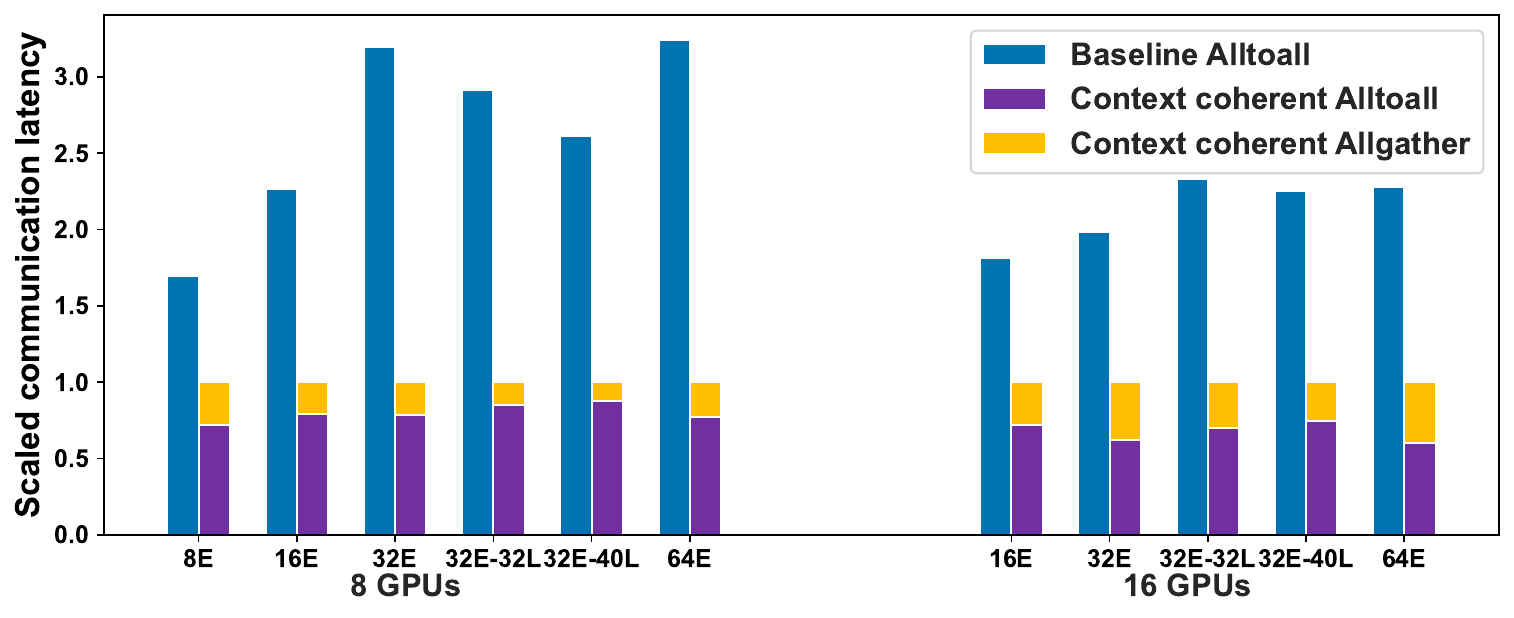}
    % \caption{Overhead in communication with context-coherent Alltoall and AllGather.}
    \label{fig:subfigureA}
  \end{subfigure}
  \\
  \vspace{-1em}
  \begin{subfigure}[b]{0.29\textwidth}
    \includegraphics[width=\linewidth]{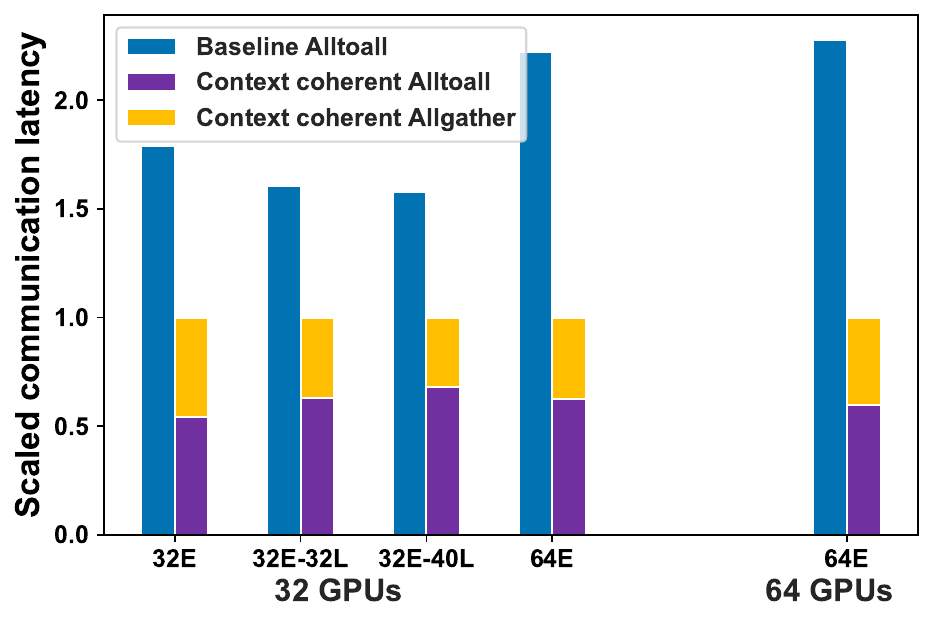}
    % \caption{Overhead in communication with context-coherent Alltoall and AllGather.}
    \label{fig:subfigureC}
  \end{subfigure}
  \vspace{-1.5em}
  \caption{On various pre-trained GPT MoE models, we change the expert parallel size and test overall communication overhead. 32L and 40L denote the GPT models with 32 and 40 layers.}
  \vspace{-1em}
  \label{fig:coherent_comm}
\end{figure}
\subsection{Reducing Collective Communications with Context-Coherent Expert Parallelism}
In the vanilla expert parallelism, each MoE layer strictly requires two Alltoall collectives because tokens need to be gathered by their corresponding GPU in the data-parallel group to perform the attention computation. With our context-coherent design, however, contexts of all tokens are coherent and visible on every GPU, meaning that we no longer need the second Alltoall to retrieve tokens, instead, they can perform in-place attention computation on any GPUs. Fig~\ref{fig:coherent_comm} shows the communication overhead in the original expert parallelism and our context-coherent design.

We tested the GPT 350M model with 8, 16, 32, and 64 experts per layer respectively, and found out that with context coherence, a large proportion of Alltoall communication becomes redundant as tokens will perform in-place attention computation. Note that the reduction in Alltoall communication overhead is more than 50\%, the reason is that tokens might find their experts on local GPUs even though these experts are not loaded in a topology-aware manner, thus they can be directly routed to those experts without going back their original GPU.
Also, we found the overhead of using AllGather to assure context coherence at the end of each iteration is trivial in 8 and 16-GPU cases. Though it becomes slightly heavier with 32 and 64 GPUs, the overall communication is still much less than the baseline. Also, as the model gets 32 and 40 layers, AllGather becomes less significant.

\subsection{Reduced Cross-GPU Token Routing with Expert Affinity}
\begin{figure}[h]
    \centering
    \vspace{-1.0em}
    \includegraphics[width=0.42\textwidth, page=1]{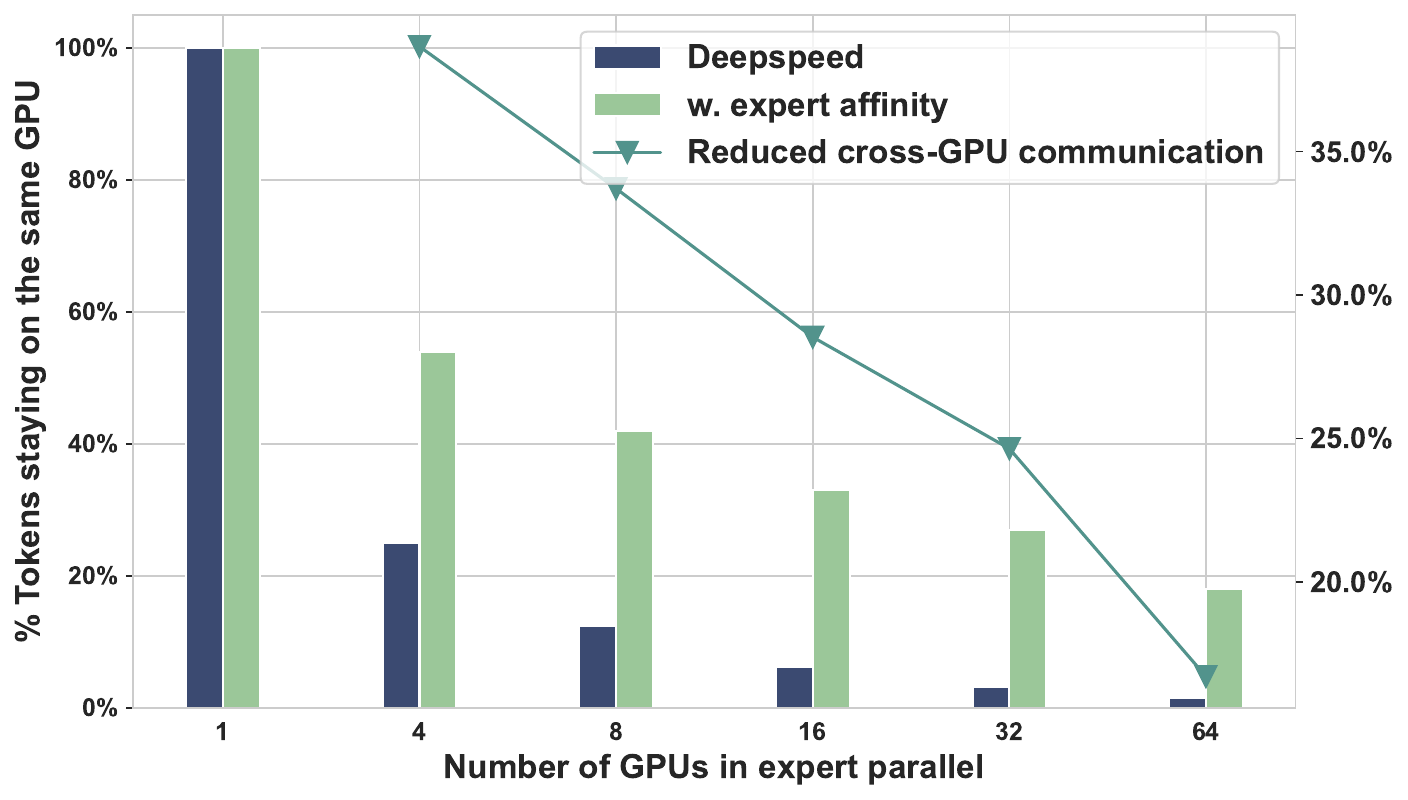}
    \caption{Evaluated on pre-trained GPT 350M MoE-64, bars denote the average percentage of tokens that are routed to experts on their current GPU. The plot shows how much cross-GPU communication is reduced with our expert affinity design.}
    \label{fig:save_gpu_comm}
\end{figure}

Fig~\ref{fig:save_gpu_comm} illustrates the reduced cross-GPU communication from two perspectives. First, given a MoE-64 model, when using less GPU to perform expert parallelism, each GPU will hold more experts per layer. For example, when using 4 GPUs, each GPU holds 16 experts per layer, compared to when using 32 GPUs, each GPU can only hold 2 experts per layer, thus a token might be more possible to be routed to other GPUs. The baseline Deepspeed framework does not have any optimization on the placement of inter-layer experts, which means that tokens can be routed to any GPU with an equal chance. In our design, since we exploit the expert affinity between layers, on 4 GPUs, we can observe an average of over half of tokens are not involved in the Alltoall communication. When scaling out to 8 GPUs, our expert affinity design can keep $40\%$ tokens remaining on the same GPU, while the baseline drops drastically. When we load the model with 32 GPUs, we can still maintain it at $28\%$. Furthermore, the plot depicts the improvement in reducing the number of outgoing tokens that will actually require Alltoall communications, where $40\%$ of communication is saved when using 4 GPUs, and $25\%$ is saved when using 32 GPUs.

\subsection{Reduced Inter-node Token Routing with Expert Affinity}
\begin{figure}[h]
    \centering
    \vspace{-1.0em}
    \includegraphics[width=0.42\textwidth, page=1]{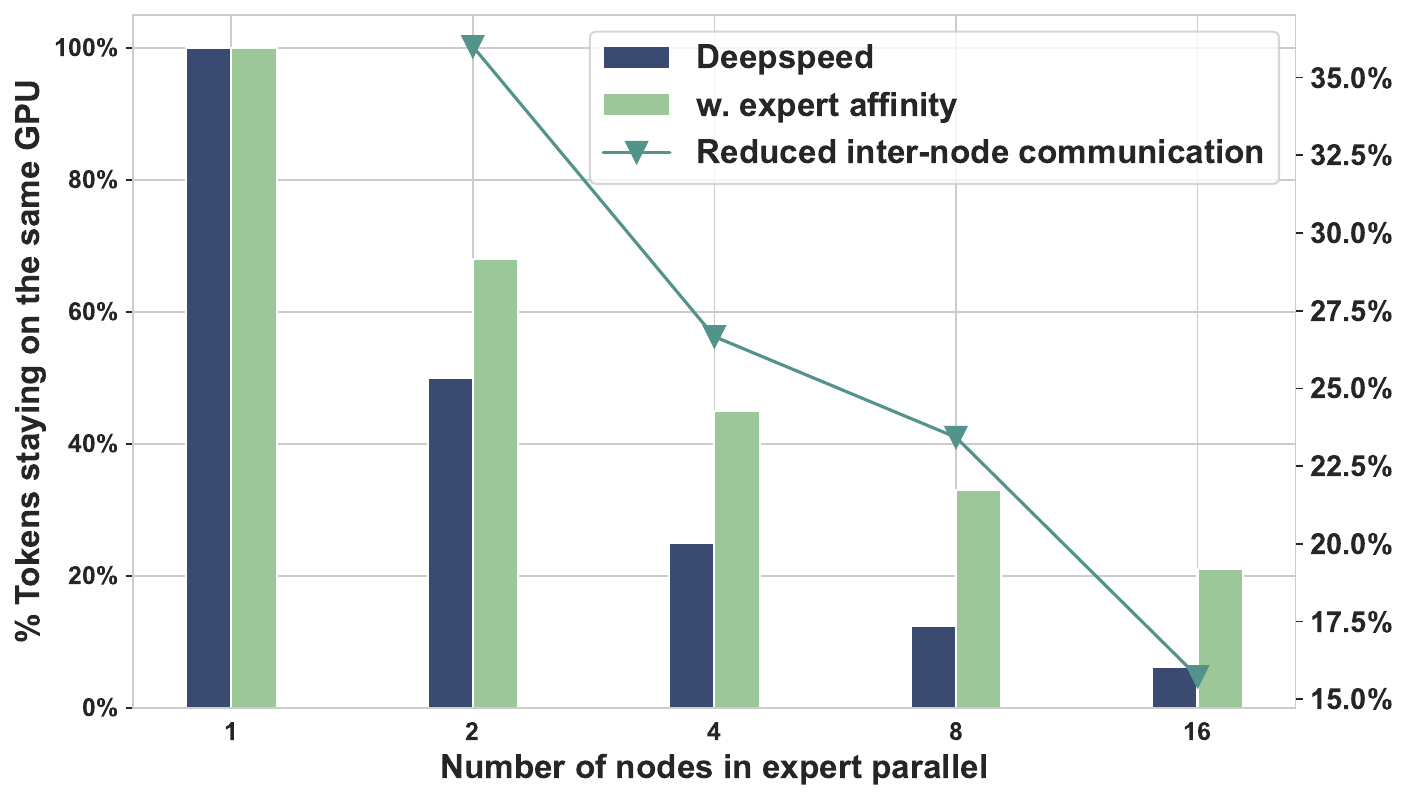}
    \caption{Similar to the previous figure, but here bars denote the average percentage of tokens that are routed to experts on their current node. The plot shows how much inter-node communication is reduced with our expert affinity design.}
    \label{fig:save_node_comm}
    \vspace{-1em}
\end{figure}
Similar trends can be observed in Fig~\ref{fig:save_node_comm}, where we measure the number of tokens that are routed to experts intra-node. As our staged expert affinity aims to reduce the inter-node routing with the highest priority, more tokens are likely to stay in the same node rather than being routed to experts on other nodes. In our experiments, we found that with the expert affinity design, tokens are average 2x more likely to stay within the same node without being involved in inter-node communication.

% \begin{figure}[!t] % Use figure* to span both columns
%     \centering
%     \begin{minipage}{0.48\textwidth}
%         \centering
%         \begin{subfigure}{\textwidth}
%             \centering
%             \includegraphics[width=0.90\linewidth]{main_result.png}
%         \end{subfigure}
        
%         \begin{subfigure}{\textwidth}
%             \centering
%             \includegraphics[width=0.90\linewidth]{main_result1.png}
%         \end{subfigure}
        
%         \begin{subfigure}{\textwidth}
%             \centering
%             \includegraphics[width=0.90\linewidth]{main_result.png}
%         \end{subfigure}
        
%         \begin{subfigure}{\textwidth}
%             \centering
%             \includegraphics[width=0.90\linewidth]{main_result.png}
%         \end{subfigure}
%     \end{minipage}
%     \hspace{0.04\textwidth} % Space between the two columns
%     \begin{minipage}{0.48\textwidth}
%         % Place other content here (text, figures, etc.)
%     \end{minipage}
%     \caption{Common caption for all the figures}
%     \label{fig:mylabel}
% \end{figure}

% GPU has >=2 capacity, Cambridge 4-GPU node, 8/16/32/64 experts
% Deepspeed, FastMoE, FasterMoE, SmartMoE, Tutel

\begin{figure}[h]
  \centering
  \begin{subfigure}[b]{0.22\columnwidth}
    \centering
    \includegraphics[width=\linewidth]{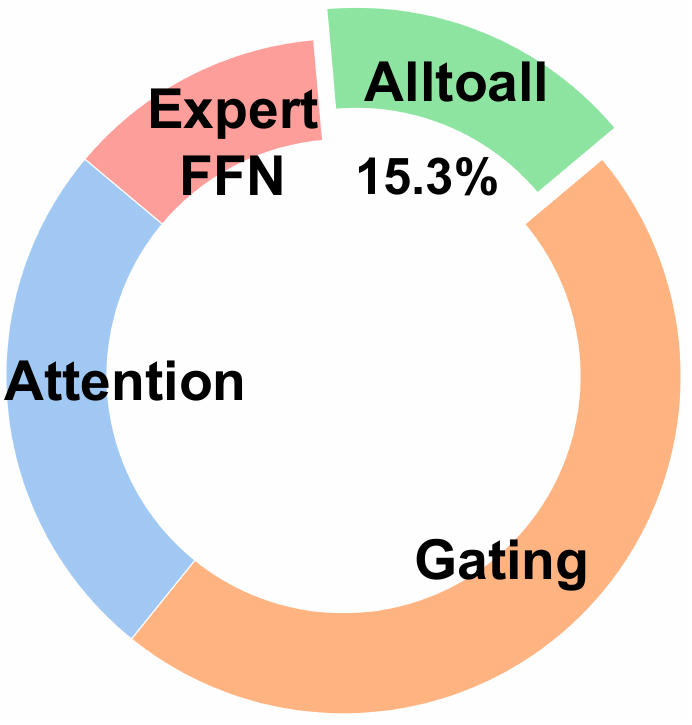}
    \caption{1 node}
    \label{fig:prop_1node}
  \end{subfigure}
  \begin{subfigure}[b]{0.22\columnwidth}
    \centering
    \includegraphics[width=\linewidth]{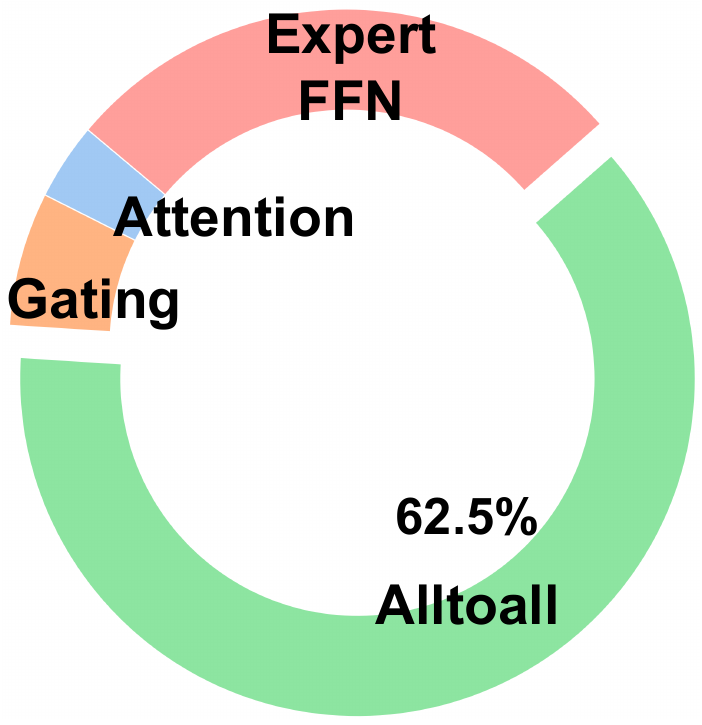}
    \caption{2 nodes}
    \label{fig:prop_2node}
  \end{subfigure}
  \begin{subfigure}[b]{0.22\columnwidth}
    \centering
    \includegraphics[width=\linewidth]{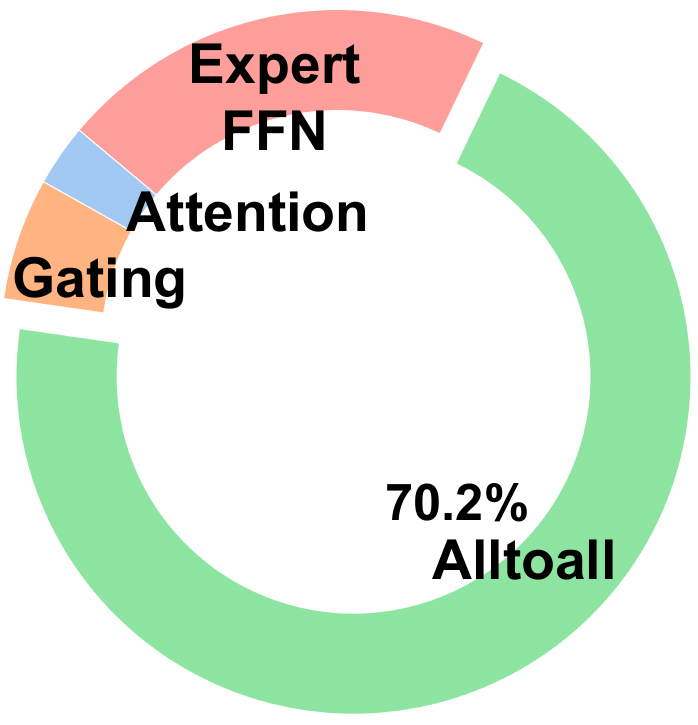}
    \caption{4 nodes}
    \label{fig:prop_4node}
  \end{subfigure}
  \begin{subfigure}[b]{0.22\columnwidth}
    \centering
    \includegraphics[width=\linewidth]{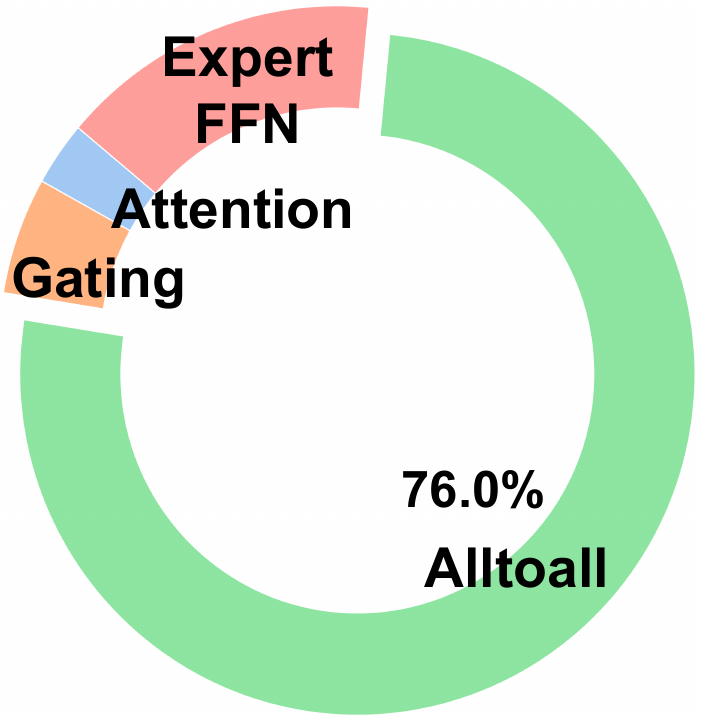}
    \caption{8 nodes}
    \label{fig:prop_8node}
  \end{subfigure}
  \caption{Proportion of Alltoall overhead to the time spent on computations. Here we only measure the most significant four operations in the MoE model, as others are trivial.}
  \label{fig:prop_nodes}
\end{figure}
\begin{figure*}[t]
  \includegraphics[width=0.97\textwidth, page=1]{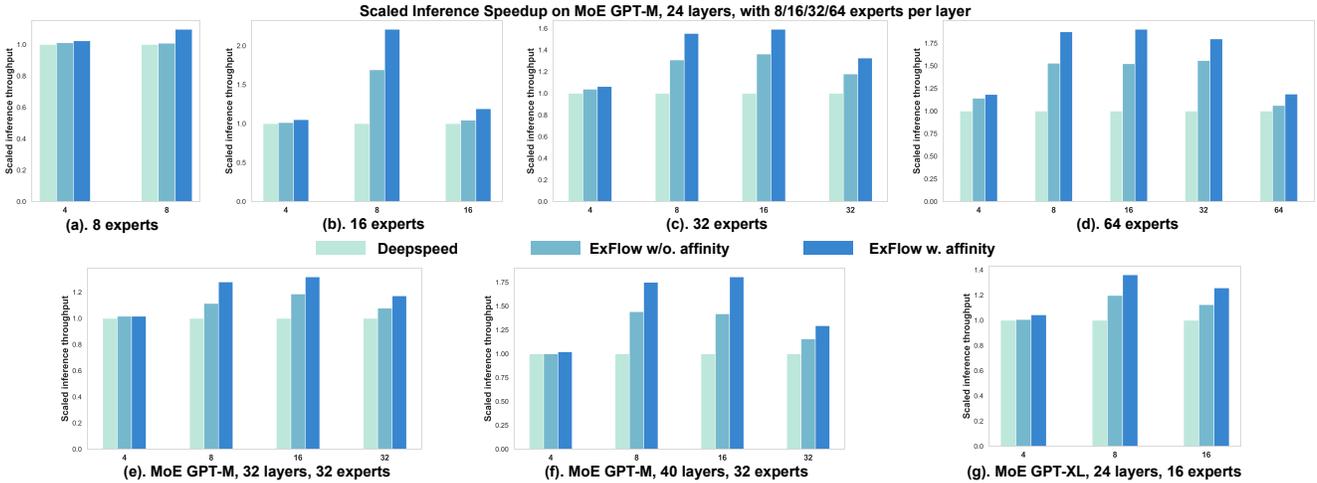}
  \caption{End-to-end GPT MoE model inference throughput. We test 7 variants of pre-trained models with multiple compute nodes, each with 4 GPUs. Results are normalized for better visualization.}
  % \vspace{-2em}
  \label{fig:main_res}
\end{figure*}

\subsection{Benchmarking GPT MoE Inference}
After examining the property of tokens' local routing, we now perform an end-to-end inference test on the entire pre-trained model. When using different numbers of nodes to perform expert parallelism, the proportion of Alltoall overhead varies greatly. Fig~\ref{fig:prop_nodes} depicts the ratio that each operation takes over in the MoE model. When using only one node, all GPUs are connected internally via NVLINK, which is of high speed and low latency, thus, the overhead of Alltoall communication is about $15\%$, and computation dominates the overall time. In this situation, there will not be too much space for us to optimize.

As we include more compute nodes, the overhead of Alltoall becomes more significant in the vanilla expert parallelism. When using 2 nodes, we observe a surge of Alltoall overhead to about $63\%$ of the overall time. When scaling out to 8 nodes, the inference is almost purely communication-bounded, with $76\%$ of time spent on Alltoall. Fig~\ref{fig:main_res} shows four different pre-trained GPT 350M MoE models under a series of parallel configurations. For MoE-8 model, we use 4 and 8 GPUs to perform expert parallel, since Alltoall overhead becomes more salient in inter-node communication, our expert affinity strategy brings $10\%$ speedup when using 8 GPUs. For MoE-16 model, we observe the most significant \textbf{2.2x} speedup is obtained when each GPU holds 2 experts per layer. When scaling out to 16 GPUs where each GPU only holds 1 expert per layer, the improvement is about $20\%$. In MoE-32 models, when the model is running on 8 and 16 GPUs, our methods can achieve \textbf{1.6x} speedup. And similarly, for MoE-64 model, the highest gains in throughput are when each GPU holds 8, 4, 2 experts per layer. 

By examining the trend in these results, we find an interesting behavior that when each GPU holds more experts, context coherence and expert affinity design can bring more performance gain because it can largely exploit the expert affinity within each GPU, meaning that it can save most Alltoall communications. However, when each GPU only holds 1 expert per layer, the expert affinity will mostly be at the intra-node level, e.g. MoE-32 on 32 GPUs, and MoE-64 on 64 GPUs. In these cases, the overhead in introducing more nodes in communication becomes salient compared to what we saved with intra-node expert affinity. For the 4-GPU case, although each GPU holds many experts per layer, there is not much performance gain due to that intra-node Alltoall overhead being trivial on the hardware system that our experiments are conducted on.

\subsection{Evolving Properties of Expert Affinity during the MoE Model Training}

\begin{figure}[t]
  \centering
  \vspace{-1.2em}
  \begin{subfigure}[b]{0.48\columnwidth}
    \centering
    \includegraphics[width=\linewidth]{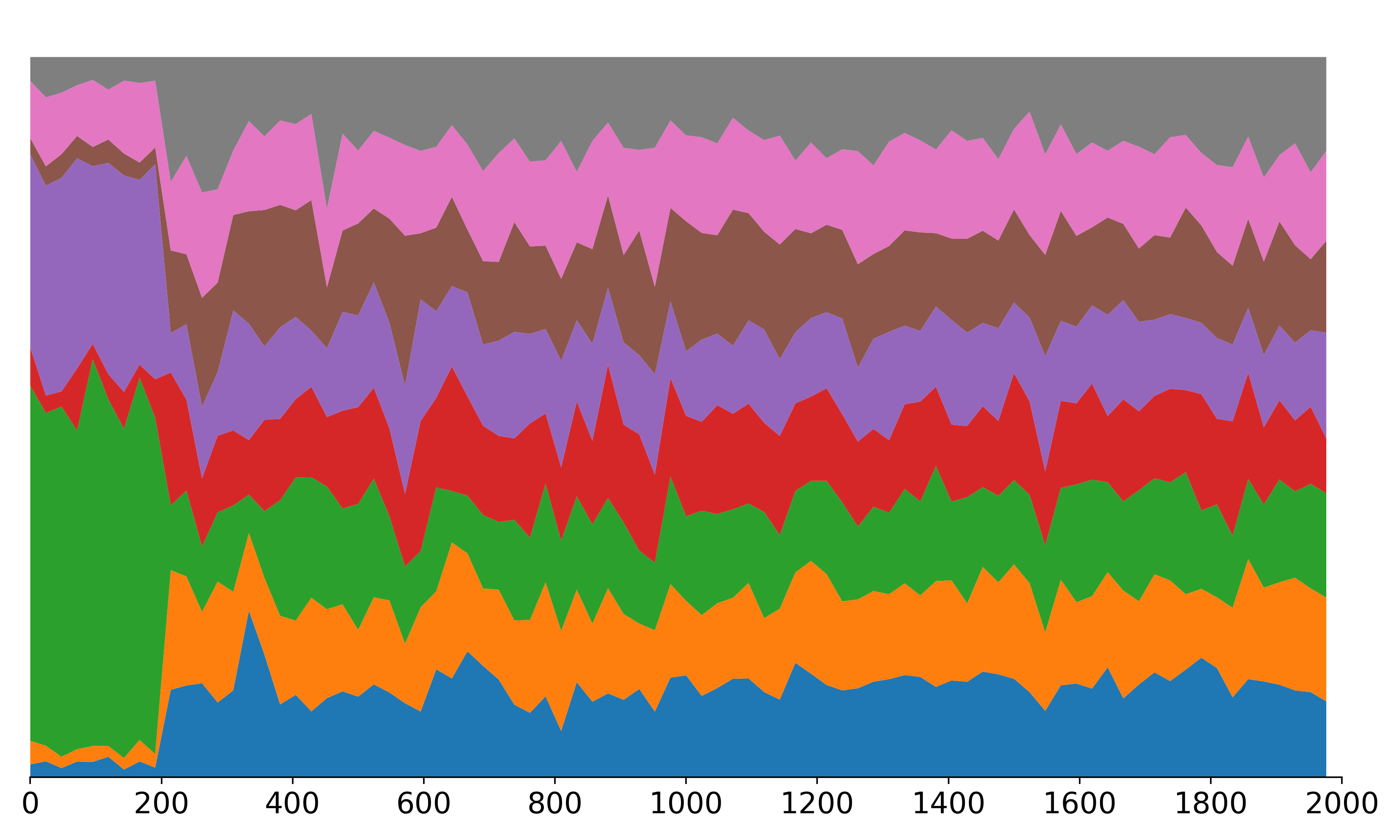}
    \caption{GPT MoE-8}
    \label{fig:ratio8_start}
  \end{subfigure}
  \begin{subfigure}[b]{0.48\columnwidth}
    \centering
    \includegraphics[width=\linewidth]{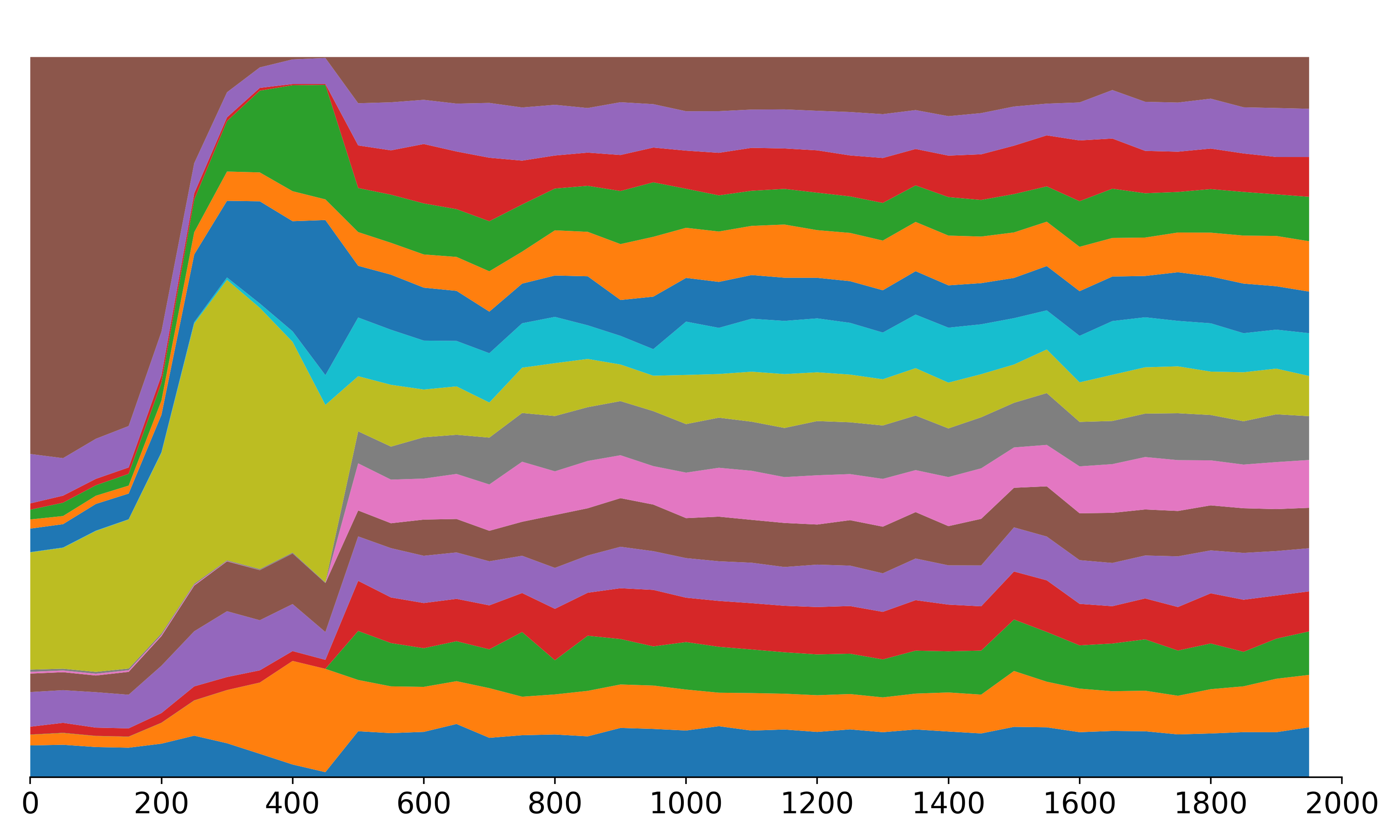}
    \caption{GPT MoE-16}
    \label{fig:ratio16_start}
  \end{subfigure}
  \\
  \begin{subfigure}[b]{0.48\columnwidth}
    \centering
    \includegraphics[width=\linewidth]{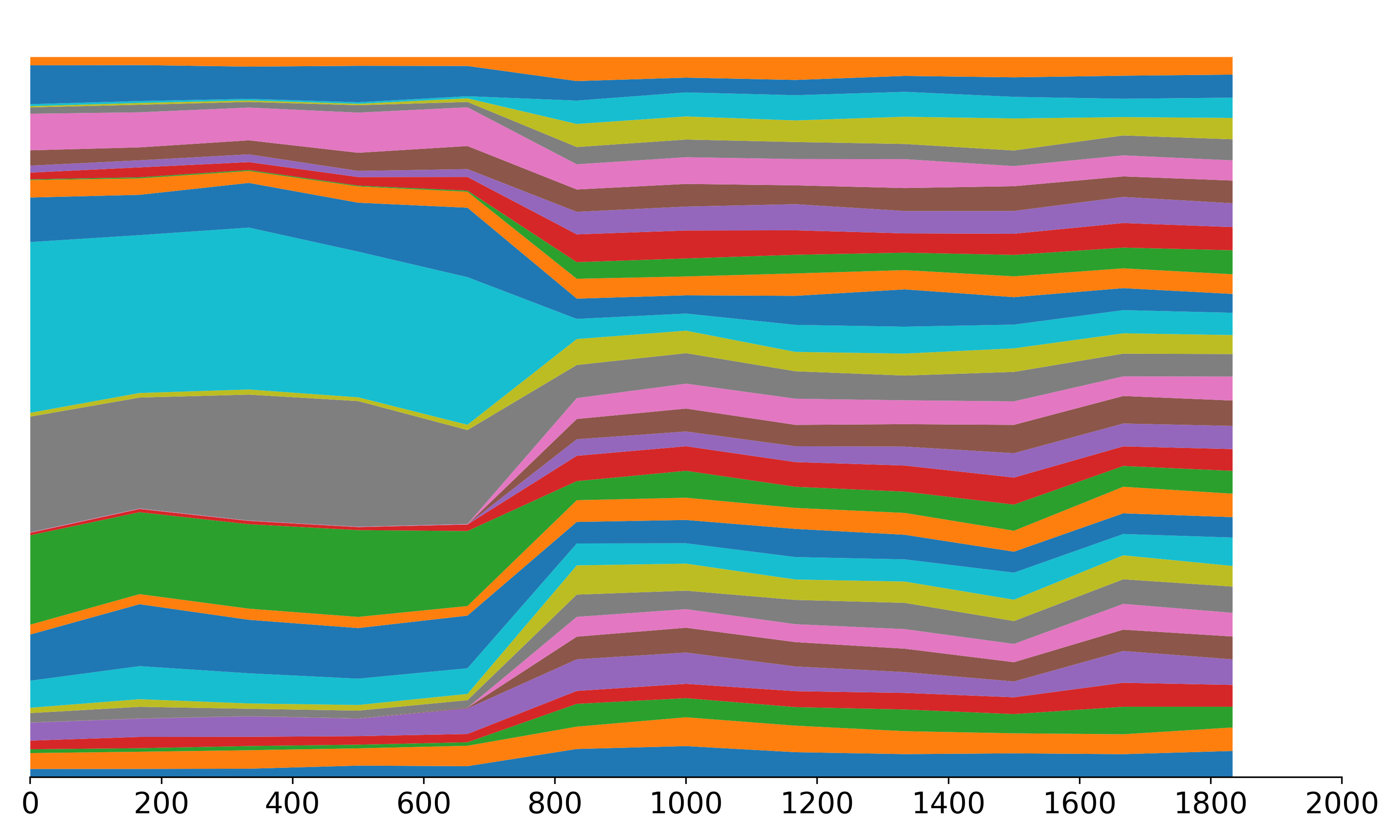}
    \caption{GPT MoE-32}
    \label{fig:ratio32_start}
  \end{subfigure}
  \begin{subfigure}[b]{0.48\columnwidth}
    \centering
    \includegraphics[width=\linewidth]{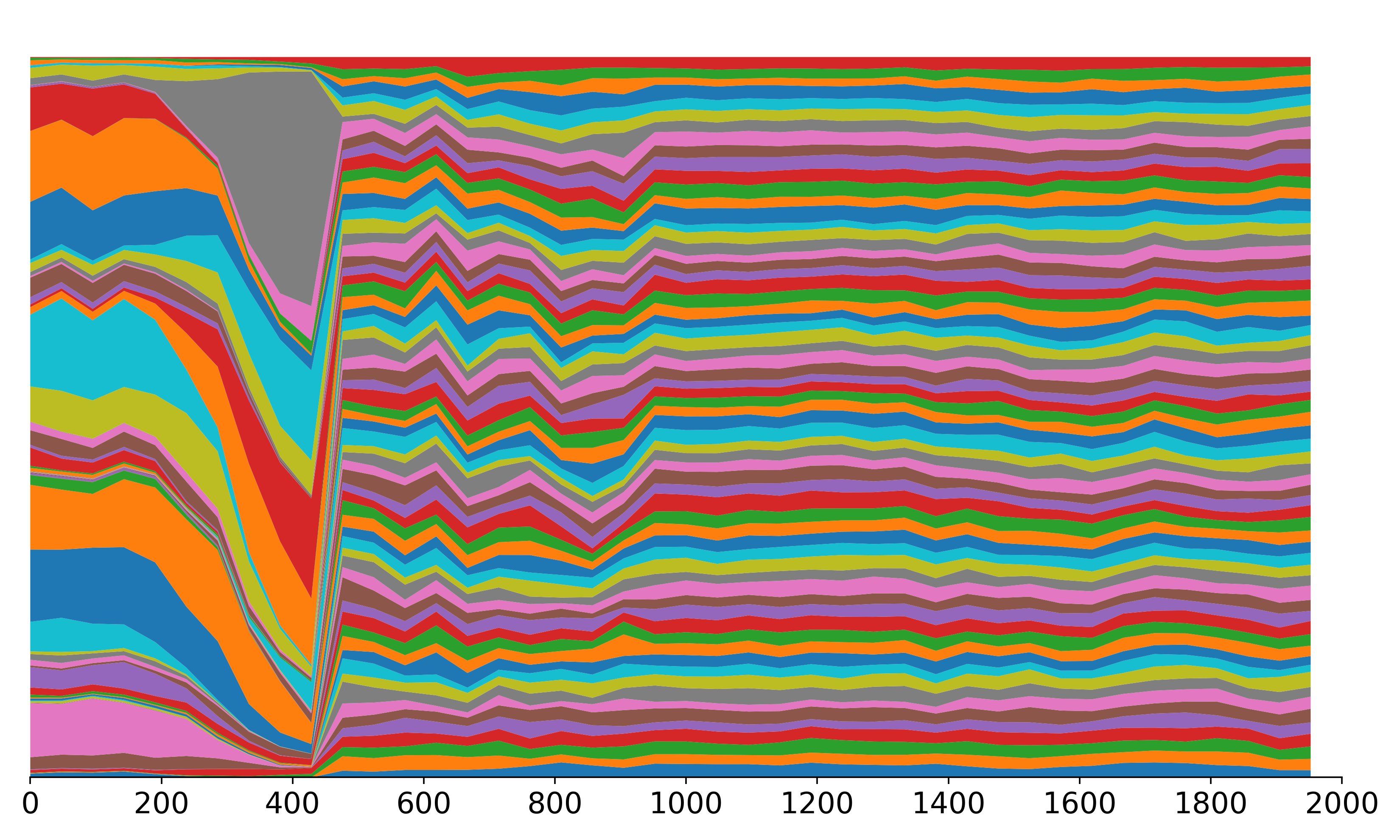}
    \caption{GPT MoE-64}
    \label{fig:ratio64_start}
  \end{subfigure}
  \caption{Proportion of tokens routed to each expert at the last MoE layer, each color represents an expert. This figure shows training iteration 0 to 2000, as training starts with random model parameters, the first hundreds of iterations see a few experts getting most of tokens. Models are trained with GShard loss, therefore, they all exhibit load balance on expert selection.}
  \label{fig:ratio_start}
  \vspace{-1.2em}
\end{figure}

\begin{figure}[h]
    \begin{subfigure}[b]{\columnwidth}
    \centering
    \includegraphics[width=0.90\linewidth]{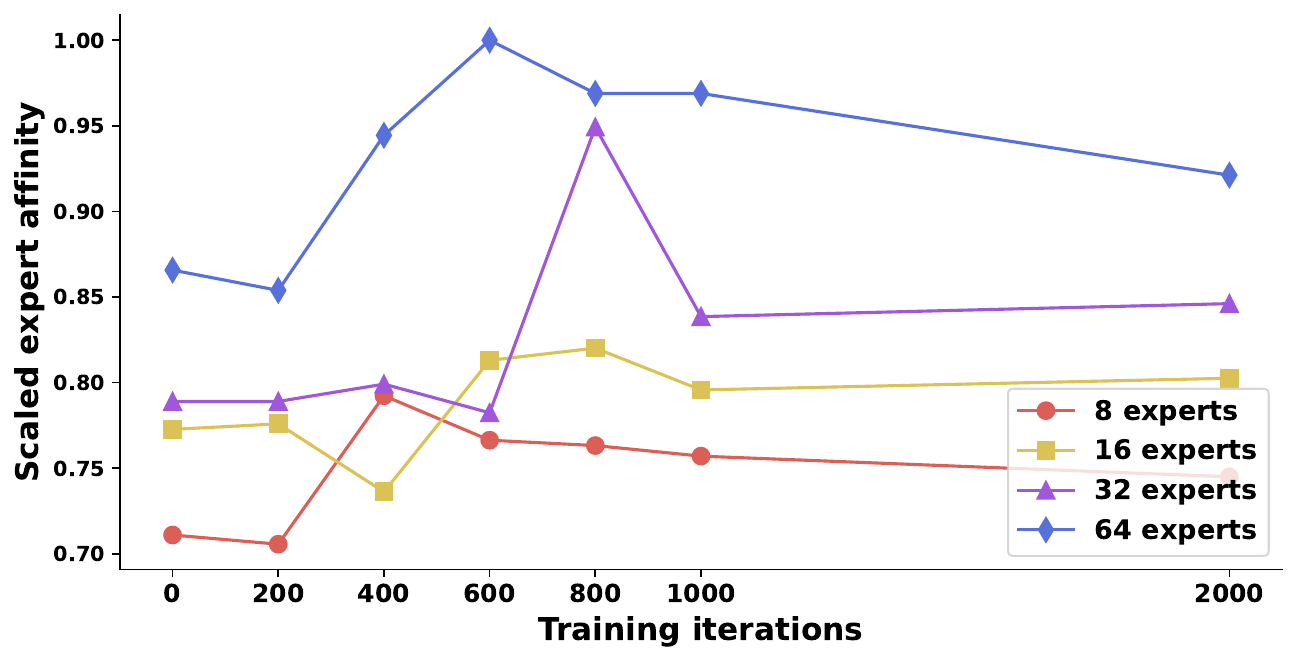}
    \caption{Scaled expert affinity during the training iteration 0 to 2000. We observe some oscillations in the beginning.}
    \label{fig:trend_start}
  \end{subfigure}
  \\
  \begin{subfigure}[b]{\columnwidth}
    \centering
    \includegraphics[width=0.90\linewidth]{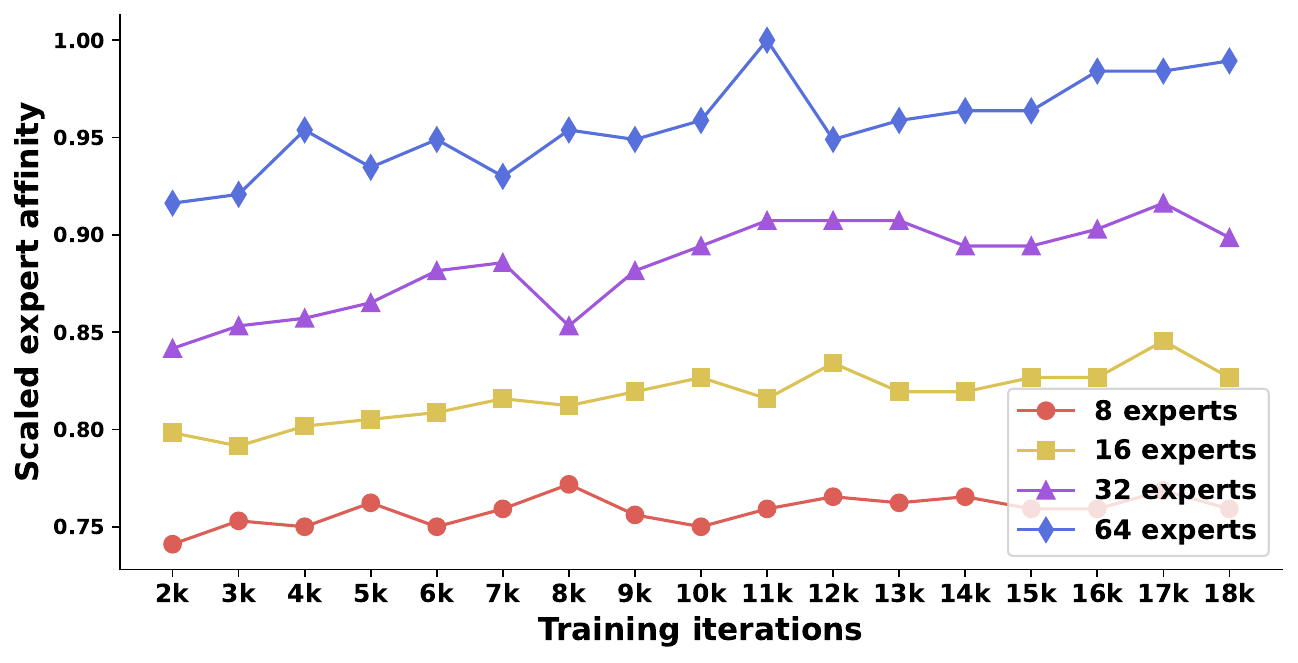}
    \caption{Scaled expert affinity during the training iteration 2000 to 18000. As the training proceeds, expert affinity steadily increases.}
    \label{fig:trend_end}
  \end{subfigure}
  \label{fig:trend}
  \vspace{-1.5em}
  \caption{Investigate the expert affinity as the model is trained from scratch. The affinity is scaled for better visualization.}
\end{figure}
In this part, we want to investigate how expert affinity evolves with the model training. Fig~\ref{fig:ratio_start} shows the expert routing proportion at the start of training. Here we only show the stats of the last MoE layer for simplicity, as we validate that other MoE layers have similar distribution. At the onset of training, the model exhibits a highly skewed distribution, indicative of a pronounced imbalance among experts, which matches the result in other studies~\cite{he2022fastermoe}. However, as the training advances, a more uniform and balanced distribution is observed. This is also reflected in Fig~\ref{fig:trend_start}, where we measure the expert affinity by solving formula~\ref{eq:obj} at different iterations of the training. The first hundreds of iterations see only a few experts frequently activated per MoE layer, and the model can indeed have a very high expert affinity because most tokens are routed to a fixed set of experts at every layer. After passing the initial stage, the expert routing distribution becomes diverse, and therefore affinity decreases as there are more experts involved in the routing. After the first 2k iterations of training, the model starts to exhibit a much more steady expert affinity and it keeps increasing as experts become more domain-specific, thus the affinity gets more salient among experts at different layers.

\subsection{How Many Tokens Are Needed to Capture the Expert Affinity in a Pre-Trained Model?}

\label{sample_tokens}
\begin{figure}[h]
    \centering
    \includegraphics[width=0.42\textwidth, page=1]{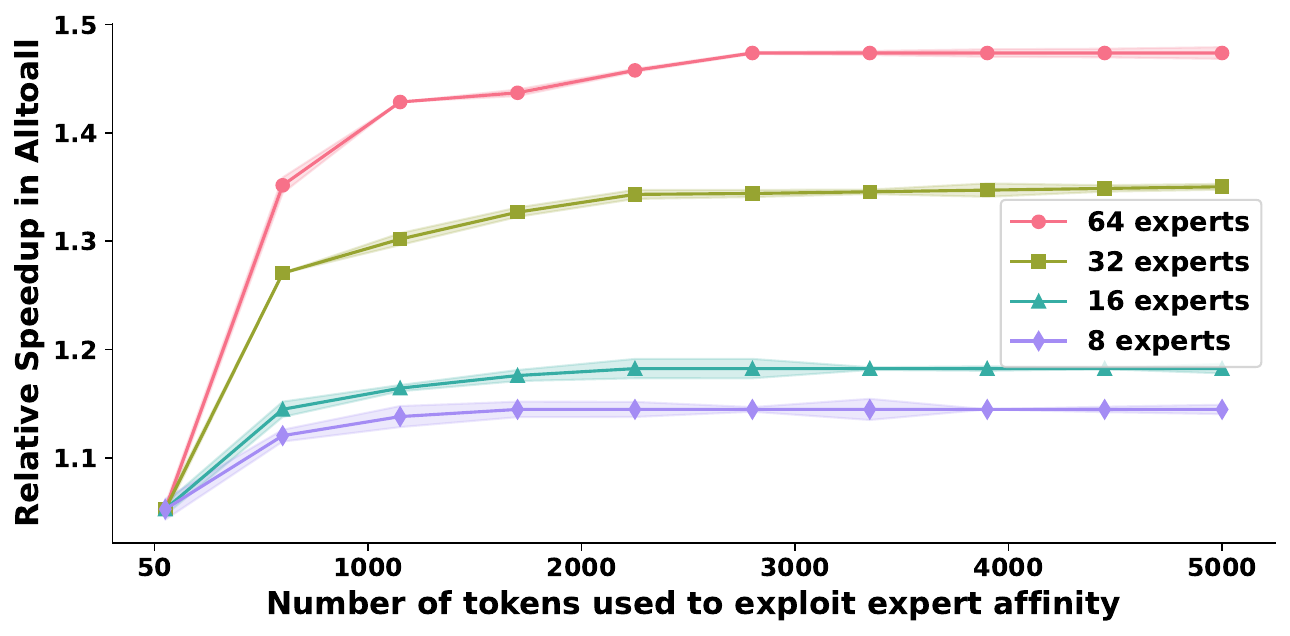}
    \caption{Number of randomly sampled tokens used to estimate the expert affinity and its relative speedup during inference. Models with more experts per layer require more tokens to precisely capture the expert affinity.}
    \label{fig:num_tokens}
\end{figure}

In \ref{dataset}, we briefly mentioned how to properly solve the integer linear programming problem~\ref{eq:obj} to get the expert affinity in a pre-trained MoE model. In practice, however, since the Pile dataset contains hundreds of billions of tokens, it is infeasible to trace and record all tokens' routing decisions. Therefore, we chose to randomly sample a portion of tokens. Fig~\ref{fig:num_tokens} shows the relative speedup in Alltoall communication when we use different numbers of tokens to capture the expert affinity. Since expert affinity is essentially a form of conditional probability among inter-layer experts' routing preferences, using more tokens' information will definitely give a better approximation. Here, we find that given the pre-trained GPT MoE models, we typically only need thousands of tokens to precisely capture the expert affinity. For MoE-8 models, 1000 tokens are enough, and for MoE-64 models, 3000 tokens are sufficient. Therefore, formula~\ref{eq:obj} can be solved efficiently by only examining these many tokens.

\subsection{Consistency on Out-of-distribution Datasets}
In deployment, MoE models might be used in scenarios where the context is different from its training dataset. Then the question is, do MoE models always exhibit similar expert affinity regardless of the context of input data? In other words, is expert affinity truly an intrinsic property of the pre-trained MoE model?  
\begin{table}[h]
\centering
% \fontfamily{phv}\selectfont
\small % Reduce font size
\renewcommand{\arraystretch}{1.5}
\begin{tabular}{c|c|c|c|c}
\toprule[1pt]
   & Pile~\cite{gao2020pile}                         & C4~\cite{c4_dataset}   &  Dolma~\cite{dolma}   & Yelp~\cite{yelp_dataset}  \\ 
\hline
Intra-GPU & 1.000 & 0.998 & 0.998 & 1.005 \\
Intra-Node & 1.000 & 0.997 & 0.989 & 1.003 \\
\hline
\end{tabular}
\caption{Using Pile to profile expert affinity and test on C4, Dolma, and Yelp. Numbers are row-normalized.}
% \fontfamily{\rmdefault}\selectfont
\renewcommand{\arraystretch}{1}
\label{tab:ood}
\vspace{-1em}
\end{table}

Table~\ref{tab:ood} shows the expert affinity on datasets that are not included in the training data. We use the Pile~\cite{gao2020pile} dataset to profile the expert affinity of a GPT 350M MoE-32 model, then we directly measure if this expert affinity holds true on three more datasets, namely, C4~\cite{c4_dataset}, Dolma~\cite{dolma}, and Yelp Reviews~\cite{yelp_dataset}. The expert placement we solved from the Pile dataset shows almost identical affinity on other out-of-distribution datasets, proving the expert affinity an inherent characteristic in pre-trained MoE models.

% \subsection{The robustness and insensitivity of expert affinity on Out-of-distribution(OOD) dataset}
% Out-of-Distribution (OOD) data pertains to data points that diverge from the distribution of the training dataset, and assessing a model’s adaptability and performance on such data is pivotal for gauging its generalization capabilities. For our proposed expert affinity, we would like to see if these experts remain closely related when we directly apply the pre-trained GPT MoE model to a new dataset. This scrutiny is crucial as it sheds light on the experts' resilience and versatility, providing a more comprehensive view understanding of expert affinity in LLMs. Also, it ensures that given a well-trained MoE model, we only need to solve out its expert affinity on its training set and it can then be applied to various datasets, making it more applicable in real-world scenarios.

%% file: 7-relatedwork.tex
\section{Related Work}
While numerous prior works exist to optimize the pre-training step of MoE models. While their methods to achieve this differ such as combining expert parallelism with other parallelism strategies like tensor parallelism \cite{singh-ted} and sharded optimizers \cite{artetxe2021efficient}, or developing optimized routing kernels \cite{nie2022hetumoe, hwang2023tutel}, or using CPU and SSD offload \cite{shen2022se}, most strategies are optimized for the pre-training MoE training paradigm and hardware. Our work is complementary to these, and is exclusively applied at inference time. 

Jiamin Li \textit{et al.} proposed the expert popularity~\cite{li2023accelerating} between two consecutive layers. However, they only calculate the top-k popular experts at every MoE layer and then create the replica of those most popular experts on local GPUs. This is similar to formula~\ref{eq:simple} in our methodology section. As we discussed, this only guarantees a local optima for the specific experts, therefore, instead of performing global expert placement optimization, they use extra memory to accommodate these popular experts locally. In our design, we do not need such explicit replicas of popular experts as we form it as a global optimization object function. In the most extreme cases, where GPU's memory can only accommodate one expert per layer, our method can still provide speedup by leveraging intra-node expert affinity, as shown in Fig~\ref{fig:main_res}. 

Jiaao He \textit{et al.} proposed FasterMoE~\cite{he2022fastermoe}, and Chang Chen \textit{et al.}~\cite{chen2022ta} introduced TA-MoE, both optimizing large-scale MoE model training via topology-aware gating strategies. However, the validity of these strategies is significantly compromised during inference due to the varying nature of hardware topologies. The divergence in hardware configurations during inference renders the topology-aware gating approach ineffective, underscoring a critical limitation of these methods in adapting to dynamic hardware environments. Mingshu Zhai\textit{et al.} proposed SmartMoE~\cite{zhai2023smartmoe}, where they investigated an offline strategy for optimized training of MoE models. It primarily revolves around the decomposition of the hybrid parallelism space into static pools, which indeed is also to solve a combinatorial optimization problem. 
\label{sec:relatedwork}

%% file: 6-contributions.tex
\section{Contributions and Conclusion}
In conclusion, we introduced ExFlow, a novel optimization technique that significantly accelerates the inference of GPT-based Mixture of Experts (MoE) models in distributed systems. By exploiting inherent inter-layer expert affinity, ExFlow eliminates a critical Alltoall communication, reducing both latency and communication overhead. Our approach leverages an integer programming model for optimal expert placement, facilitating up to a 67\% reduction in cross-GPU routing latency and a throughput improvement of up to 120\% over existing methods, without sacrificing model accuracy. These advancements not only provide a scalable solution for MoE-based inference but also offer valuable insights into the early-stage acquisition and stabilization of expert affinity in model training, thus paving the way for future research in this domain.
\label{sec:contributions}

%% file: appendix.tex
\clearpage
\onecolumn
\appendix
\section{Extend Expert Affinity to Cross Multiple Layers}
\begin{figure*}[ht]
    \centering
    \includegraphics[width=0.9\textwidth, page=1]{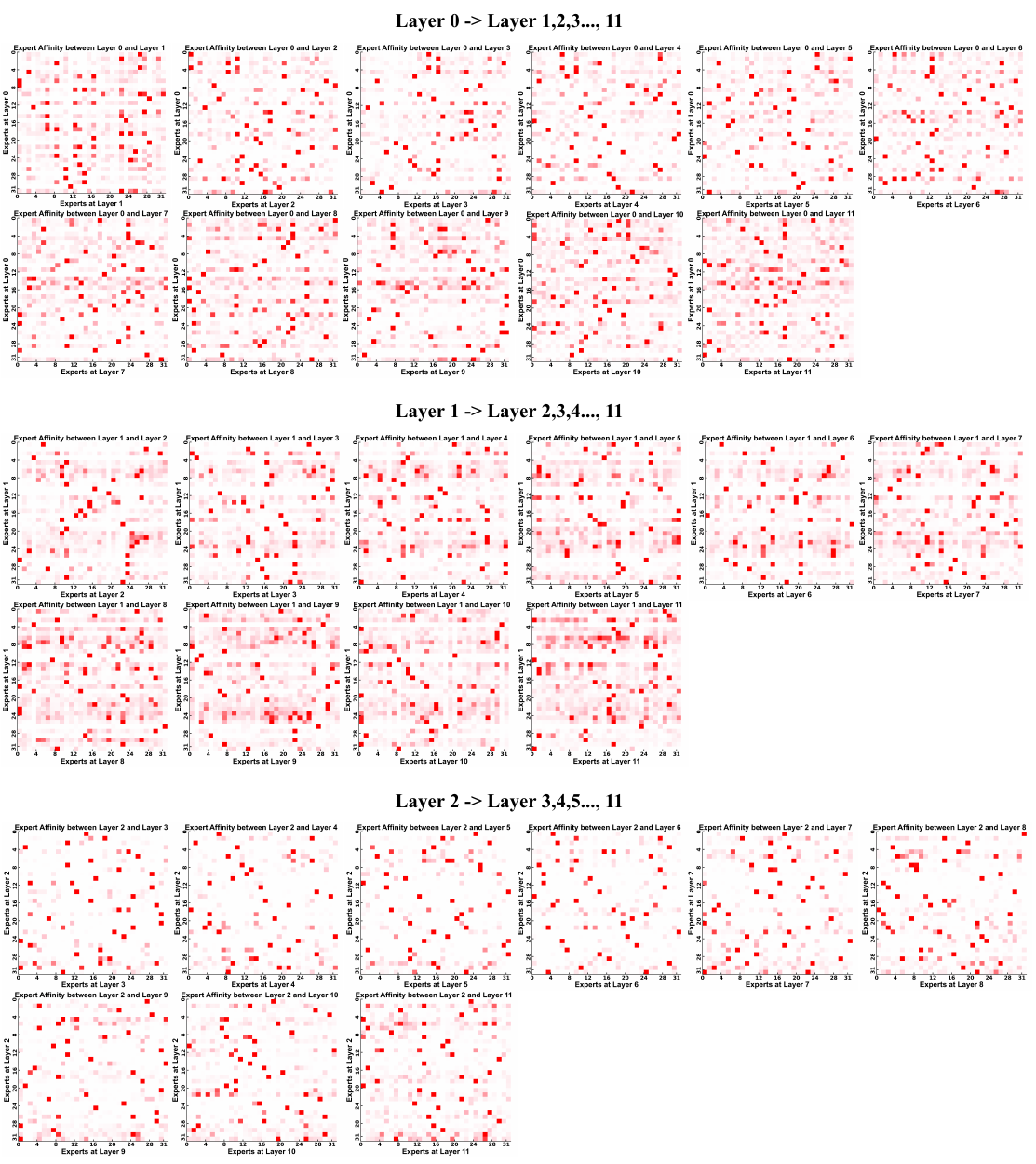}
    \caption{Pre-trained GPT 350M MoE-32 model, layer 0, 1, 2's expert affinity to all following layers.}
    \label{fig:compare}
    \vspace{-1.5ex}
\end{figure*}
\label{appendix:A}

\begin{figure*}[t]
    \centering
    \includegraphics[width=0.9\textwidth, page=1]{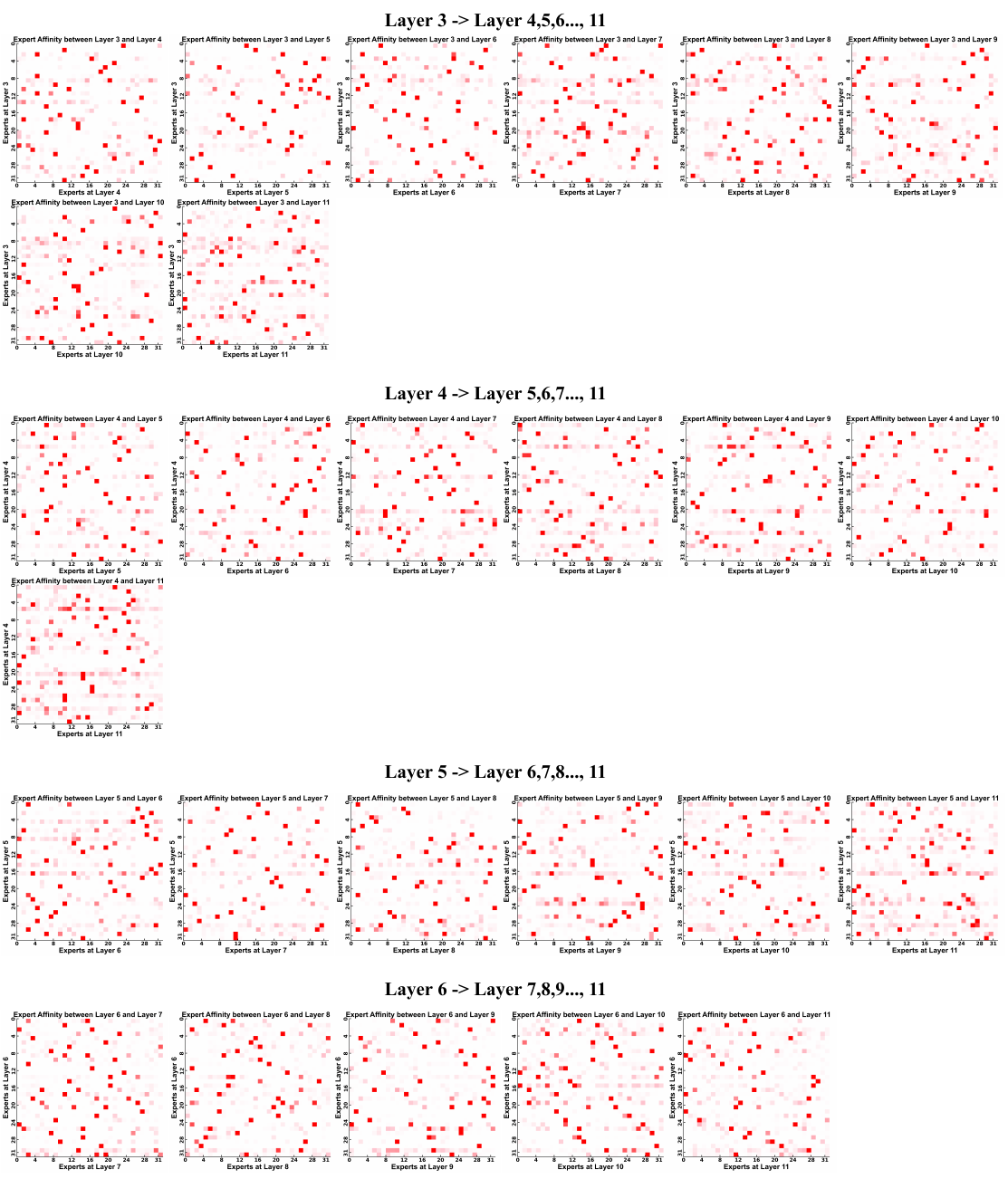}
    \caption{Pre-trained GPT 350M MoE-32 model, layer 3, 4, 5, 6's expert affinity to all following layers.}
    \label{fig:compare}
    \vspace{-1.5ex}
\end{figure*}
\label{appendix:B}

\begin{figure*}[t]
    \centering
    \vspace{-35ex}
    \includegraphics[width=0.9\textwidth, page=1]{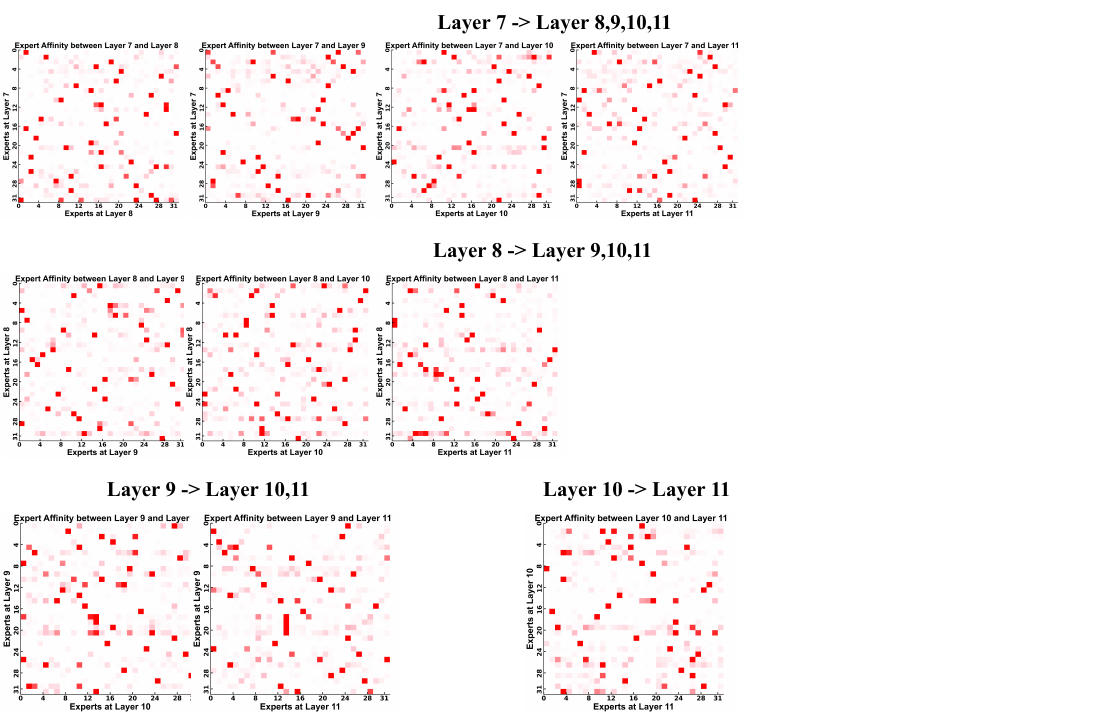}
    \caption{Pre-trained GPT 350M MoE-32 model, layer 7, 8, 9, 10's expert affinity to all following layers.}
    \label{fig:compare}
    \vspace{-1.5ex}
\end{figure*}
\label{appendix:C}